\makeatletter\def\graphicscache@inhibit{true}\makeatother

\documentclass[10pt,twocolumn,letterpaper]{article}

\usepackage[pagenumbers]{cvpr} %

\usepackage{graphicx}
\usepackage{amsmath}
\usepackage{amssymb}
\usepackage{booktabs}
\makeatletter
\@namedef{ver@everyshi.sty}{}
\usepackage{graphicscache}
\makeatother

\usepackage{placeins} %
\usepackage{balance}

\usepackage[pagebackref,breaklinks,colorlinks]{hyperref}

\usepackage[capitalize]{cleveref}
\crefname{section}{Sec.}{Secs.}
\Crefname{section}{Section}{Sections}
\Crefname{table}{Table}{Tables}
\crefname{table}{Tab.}{Tabs.}

\makeatletter
\@namedef{ver@everyshi.sty}{}
\makeatother
\usepackage{tikz}
\usepackage{pgfplots}
\pgfplotsset{
	/pgfplots/my legend/.style={
		legend image code/.code={
			\draw[thick,black](-0.05cm,0cm) -- (0.3cm,0cm);%
		}
	}
}
\pgfplotsset{compat = 1.3} %
\usepackage{threeparttable}
\usepackage{wrapfig}
\usepackage{adjustbox}
\usepackage{bm}
\usepackage{lipsum}
\usepackage{tikz, tikz-3dplot}
\usetikzlibrary{calc,arrows,arrows.meta,backgrounds,tikzmark,shapes.geometric, decorations.pathreplacing,decorations.markings, spy,matrix,positioning}
\usepackage{siunitx}
\usepackage[nolist,nohyperlinks]{acronym}
\usepackage{xcolor,colortbl} %
\usepackage{mathtools,xparse}
\usepackage{xfrac}
\usepackage{float}
\usepackage{makecell} %

\usepackage{soul}

\clearpage{}%

\newcommand{\IGNORE}[1]{}
\definecolor{ph-purple}{RGB}{129, 39, 232}
\definecolor{ph-blue}{RGB}{5, 131, 227}
\definecolor{ph-gray}{rgb}{0.5, 0.5, 0.5}
\definecolor{ph-orange}{RGB}{227, 127, 5}
\definecolor{ph-green}{RGB}{0, 135, 124}
\definecolor{ph-yellow}{RGB}{235, 201, 52}
\definecolor{ph-light-green}{RGB}{181, 209, 21}
\definecolor{ph-red}{RGB}{250, 101, 60}
\definecolor{ph-gold}{RGB}{255, 223, 0}
\definecolor{ph-metallic-gold}{RGB}{212, 175, 55}
\definecolor{ph-silver}{RGB}{192, 192, 192}
\colorlet{g}{ph-gold!90!black}
\colorlet{s}{ph-silver!60}
\colorlet{ph-orange-light}{ph-orange!70}
\colorlet{ph-blue-light}{ph-blue!70}
\colorlet{ph-purple-light}{ph-purple!70}
\colorlet{ph-green-light}{ph-green!70}
\definecolor{ph-light-gray}{rgb}{0.75, 0.75, 0.75}

\newcommand{\reffig}[1]{Fig.~\ref{#1}}
\newcommand{\reftab}[1]{Tab.~\ref{#1}}
\newcommand{\refsec}[1]{Sec.~\ref{#1}}
\newcommand{\refequ}[1]{Eq.~\ref{#1}}

\DeclarePairedDelimiter{\abs}{\lvert}{\rvert}
\DeclarePairedDelimiter{\norm}{\lVert}{\rVert}

\newcommand{\enc}{\mathrm{enc}}
\newcommand{\encOut}{\mathbf{h}}

\newcommand{\entriesPerLevel}{T}

\newcommand{\featuresPerEntry}{F}

\newcommand{\levels}{L}

\newcommand{\BigO}{\mathcal{O}}

\newcommand{\pos}{\mathbf{x}}

\newcommand{\normal}{\mathbf{n}}

\newcommand{\Real}{\mathbb{R}}

\newcommand{\LatticeParams}{\theta}
\newcommand{\NetParams}{\Phi}

\newcommand{\ViewDir}{\mathbf{v}}
\newcommand{\CamOrig}{\mathbf{o}}
\newcommand{\Pixel}{\mathbf{p}}

\newcommand{\GeomFeat}{\bm{\chi}}
\newcommand{\SigmoidSlope}{a}
\newcommand{\Loss}{\mathcal{L}}
\newcommand{\tangent}{\bm{\eta}}
\newcommand{\RandomVec}{\bm{\tau}}

\newcommand{\lipc}{k}

\newcommand{\softplus}{\mathrm{softplus}\,}

\clearpage{}%
\def\cvprPaperID{9176} %
\def\confName{CVPR}
\def\confYear{2023}

\begin{document}

\title{PermutoSDF: Fast Multi-View Reconstruction with\\
Implicit Surfaces using Permutohedral Lattices}

\author{Radu Alexandru Rosu \qquad Sven Behnke \\
University of Bonn, Germany\\
{\tt\small \{rosu, behnke\}@ais.uni-bonn.de}
}
\maketitle

\begin{acronym}
	\acro{MLP}{Multi-Layer Perceptron}
	\acro{SDF}{Signed Distance Function}
\end{acronym}

\begin{abstract}

Neural radiance-density field methods have become increasingly popular for the task of novel-view rendering. Their recent extension to hash-based positional encoding ensures fast training and inference with visually pleasing results. However, density-based methods struggle with recovering accurate surface geometry.  
Hybrid methods alleviate this issue by optimizing the density based on an underlying SDF. 
However, current SDF methods are overly smooth and miss fine geometric details. 
In this work, we combine the strengths of these two lines of work in a novel hash-based implicit surface representation. We propose improvements to the two areas by replacing the voxel hash encoding with a permutohedral lattice which optimizes faster, especially for higher dimensions. We additionally propose a regularization scheme which is crucial for recovering high-frequency geometric detail. We evaluate our method on multiple datasets and show that we can recover geometric detail at the level of pores and wrinkles while using only RGB images for supervision. Furthermore, using sphere tracing we can render novel views at 30 fps on an RTX 3090.
Code is publicly available at \url{https://radualexandru.github.io/permuto_sdf}

 \end{abstract}

\section{Introduction}

\begin{figure}
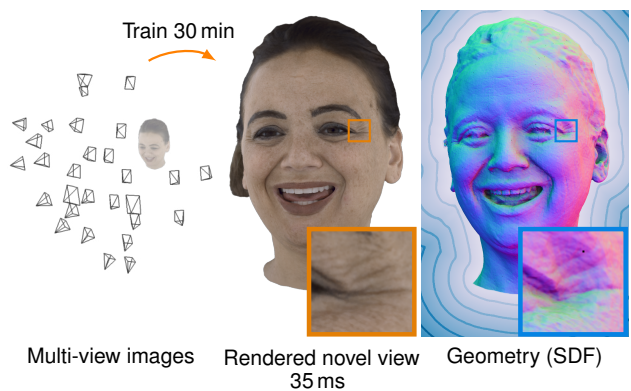

\centering
\small\sffamily
\begin{tikzpicture} [spy using outlines={ size=5cm,   every spy on node/.append style={thick}     }]

\newcommand\ShiftX{2.75}

\node[inner sep=0pt] (Mug) at (0,0)
{\includegraphics[width=.33\columnwidth]{./imgs/teaser/frustums2_composed.png}};
\node[inner sep=0pt] (App) at (\ShiftX,0)
{\includegraphics[width=.33\columnwidth,compress=false]{./imgs/teaser_v2/rgb_transparent_cropped.png}};
\node[inner sep=0pt] (Geom) at (\ShiftX*2,0)
{\includegraphics[width=.33\columnwidth,compress=false]{./imgs/teaser_v2/isolines_cropped.png}};

\spy [ph-orange,draw,height=1.4cm,width=1.4cm,magnification=5] on ($(App.center) + (0.55, 0.57)$) in node [line width=0.6mm, anchor=south east] at ($(App.south east) + (-0.1, 0.04) $);
\spy [ph-blue,draw,height=1.4cm,width=1.4cm,magnification=5] on ($(Geom.center) + (0.55, 0.57)$) in node [line width=0.6mm, anchor=south east] at ($(Geom.south east) + (-0.04, 0.04) $);

\draw[-latex,orange, line width=0.3mm] (0.5, 1.5) arc
[
start angle=130,
end angle=50,
x radius=0.7cm,
y radius =0.5cm
] ;
\node at (0.9, 1.9) {\footnotesize Train \SI{30}{\minute} };

\node at (0.0, -2.45) {\footnotesize  Multi-view images};
\node at (\ShiftX+0.05,-2.45) {\footnotesize  Rendered novel view };
\node at (\ShiftX,-2.75) {\footnotesize  {\SI{35}{\milli\second}}   };
\node at (\ShiftX*2,-2.45) {\footnotesize Geometry (SDF) };

\end{tikzpicture}
\vspace{-2mm}
\caption{ Given multi-view images, we recover both high quality geometry as an implicit SDF and appearance which can be rendered in real-time. } \label{fig:teaser}
\vspace{-3mm}
\end{figure} 

Accurate reconstruction geometry and appearance of scenes is an important component of many computer vision tasks~\cite{colmap,kinectfusion,nerf,elasticfusion}. Recent Neural Radiance Field (NeRF)-like models~\cite{nerf,ingp,mipnerf,blocknerf} represent the scene as a density and radiance field and, when supervised with enough input images, can render photorealistic novel views.

Works like INGP~\cite{ingp} further improve on NeRF by using a hash-based positional encoding which results in fast training and visually pleasing results.
However, despite the photorealistic renderings, the reconstructed scene geometry can deviate severally from the ground-truth. For example, objects with high specularity or view-dependent effects are often reconstructed as a cloud of low density; untextured regions can have arbitrary density in the reconstruction.

Another line of recent methods tackles the issue of incorrect geometry by representing the surfaces of objects as binary occupancy~\cite{unisurf} or \ac{SDF}~\cite{neus,volsdf}. This representation can also be optimized with volumetric rendering techniques that are supervised with RGB images.
However, parametrization of the SDF as a single fully-connected \ac{MLP} often leads to overly smooth geometry and color.

In this work, we propose PermutoSDF, a method that combines the strengths of hash-based encodings and implicit surfaces. We represent the scene as an SDF and a color field, which we render using unbiased volumetric integration~\cite{neus}. A naive combination of these two methods would fail to reconstruct accurate surfaces however, since it lacks any inductive bias for the ambiguous cases like specular or untextured surfaces. Attempting to regularize the SDF with a total variation loss or a curvature loss will produce a smoother geometry at the expense of losing smaller details.
In this work, we propose a regularization scheme that ensures both smooth geometry where needed and also reconstruction of fine details like pores and wrinkles. Furthermore, we improve upon the voxel hashing method of INGP by proposing permutohedral lattice hashing. The number of vertices per simplex (triangle, tetrahedron, \ldots) in this data structure scales linearly with dimensionality instead of exponentially as in the hyper-cubical voxel case. 
We show that the permutohedral lattice performs better than voxels for 3D reconstruction and 4D background estimation. 

\noindent
In summary our main contributions are:
\begin{itemize}
\setlength\itemsep{0.01em}
\item a novel framework for optimizing neural implicit surfaces based on hash-encoding,
\item an extension of hash encoding to a permutohedral lattice which scales linearly with the input dimensions and allows for faster optimization, and
\item a regularization scheme that allows to recover accurate SDF geometry with a level of detail at the scale of pores and wrinkles.
\end{itemize}

\section{Related Work}

\subsection{Classical Multi-View Reconstruction}
Multi-view 3D reconstruction has been studied for decades. The classical methods can be categorized as either depth map-based~\cite{colmap,pmvs,patchmatch,gipuma} or volume-based~\cite{vrip,kinectfusion,voxelhashing,elasticfusion}. Depth map methods like COLMAP~\cite{colmap} reconstruct a depth map for each input view by matching photometrically consistent patches. The depth maps are fused to a global 3D point cloud and a watertight surface is recovered using Poisson Reconstruction~\cite{poisson}. While COLMAP can give good results in most scenarios, it yields suboptimal results on non-Lambertian surfaces.
Volume-based approaches fuse the depth maps into a volumetric structure (usually a Truncated Signed Distance Function) from which an explicit surface can be recovered via the marching cubes algorithm~\cite{marchingcubes}. Volumetric methods work well when fusing multiple noisy depth maps but struggle with reconstructing thin surfaces and fine details.

\subsection{NeRF Models}
A recent paradigm shift in 3D scene reconstruction has been the introduction of NeRF~\cite{nerf}. NeRFs represent the scene as density and radiance fields, parameterized by a \ac{MLP}. Volumetric rendering is used to train them to match posed RGB images. This yields highly photorealistic renderings with view-dependent effects. However, the long training time of the original NeRF prompted a series of subsequent works to address this issue. 

\subsection{Accelerating NeRF}
Two main areas were identified as problematic: the large number of ray samples that traverse empty space and the requirement to query a large \ac{MLP} for each individual sample.

Neural Sparse Voxel Fields~\cite{nsvf} uses an octree to model only the occupied space and restricts samples to be generated only inside the occupied voxels. Features from the voxels are interpolated and a shallow MLP predicts color and density. This achieves significant speedups but requires complicated pruning and updating of the octree structure.

DVGO~\cite{dvgo} similarly models the scene with local features which are stored in a dense grid that  is decoded by an MLP into view-dependent color.
Plenoxels~\cite{plenoxels} completely removes any MLP and instead stores opacity and spherical harmonics~(SH) coefficients at sparse voxel positions.

Instant Neural Graphics Primitives (INGP)~\cite{ingp} proposes a hash-based encoding in which ray samples trilinearly interpolate features between eight positions from a hash map. A shallow MLP implemented as a fully fused CUDA kernel predicts color and density. Using a hash map for encoding has the advantage of not requiring complicated mechanisms for pruning or updating like in the case of octrees.

In our work, we improve upon INGP by proposing a novel permutohedral lattice-based hash encoding, which is better suited for interpolating in high-dimensional spaces. We use our new encoding to reconstruct accurate 3D surfaces and model background as a 4D space.

\subsection{Implicit Representation}
Other works have focused on reconstructing the scene geometry using implicit surfaces.
SDFDiff~\cite{sdfdiff} discretizes SDF values on a dense grid and by defining a differentiable shading function can optimize the underlying geometry. However, their approach can neither recover arbitrary color values nor can it scale to higher-resolution geometry. 

IDR~\cite{idr} and DVR~\cite{dvr} represent the scene as SDF and occupancy map, respectively, and by using differentiable rendering can recover high-frequency geometry and color. However, both methods require 2D mask supervision for training which is not easy to obtain in practice.

In order to remove the requirement of mask supervision, UNISURF~\cite{unisurf} optimizes an binary occupancy function through volumetric rendering. VolSDF~\cite{volsdf} extends this idea to SDFs. NeuS~\cite{neus} analyzes the biases caused by using volumetric rendering for optimizing SDFs and introduces an unbiased and occlusion-aware weighting scheme which allows to recover more accurate surfaces.

In this work, we reconstruct a scene as SDF and color field without using any mask supervision. We model the scene using locally hashed features in order to recover finer detail than previous works. We also propose several regularizations which help to recover geometric detail at the level of pores and wrinkles.

\section{Method Overview}

\begin{figure*}[t]
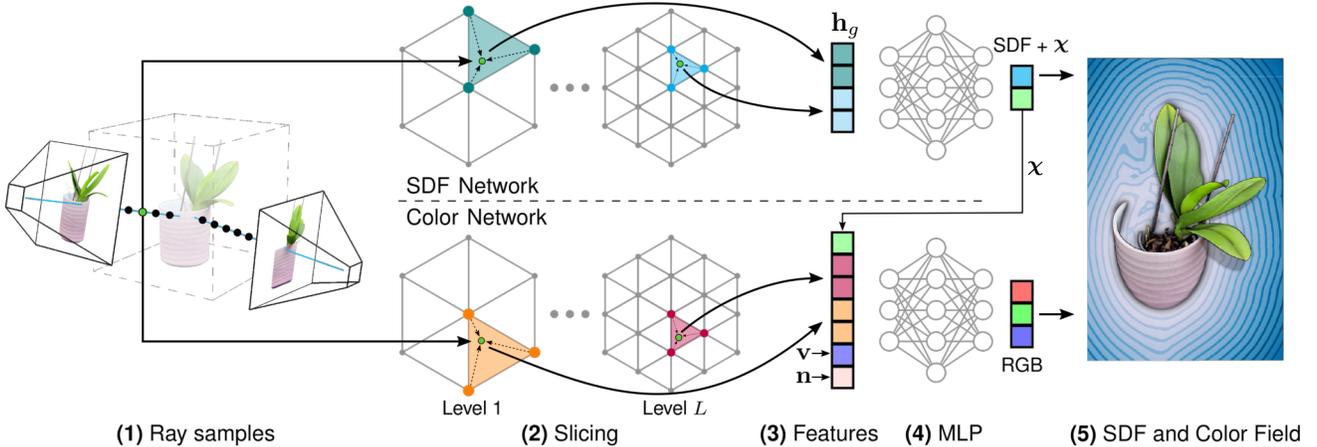

	\begin{tikzpicture}
	\begin{scope}
		\def\solidImg{./imgs/overview2/drawing6_crop.png} 
		\newlength{\WO}
		\newlength{\HO}
		\settowidth{\WO}{\includegraphics{\solidImg}}
		\settoheight{\HO}{\includegraphics{\solidImg}}
		\node[inner sep=0pt] (russell) at (9.5,0) {\includegraphics[ width=1.0\textwidth]{\solidImg}};
		\node[] at (14.6cm, 0.7cm) {\footnotesize $\GeomFeat$   };
		\fill [white] (11.7,2.4) rectangle (12.4,2.7);
		\node[] at (12.08, 2.6) { $\encOut_g$};
		
	\end{scope}
	\end{tikzpicture}
	\vspace{-5mm}
	\caption{
		Overview of our PermutoSDF pipeline. \textbf{(1)} For a batch of pixels from the posed images, we sample rays inside the volume of interest. \textbf{(2)} For each sample, we slice features from a multi-resolution permutohedral lattice. \textbf{(3)} The features from all lattice levels are concatenated. For the color network, we also concatenate additional features regarding normal $\normal$ of the SDF, view direction $\ViewDir$, and learnable features $\chi$ from the SDF network.  \textbf{(4)} Small MLPs decode the SDF and a view-dependent RGB color.  \textbf{(5)} The output is rendered volumetrically and supervised only with RGB images. We visualize surface color and a 2D slice of the SDF.}
	\label{fig:overview}
	\vspace{-2mm}
\end{figure*}

Given a series of images with poses $\{ \mathcal{I}_k \}$, our task is to recover both surface $\mathcal{S}$ and appearance of the objects within. 
We define the surface $\mathcal{S}$ as the zero level set of an SDF:
\begin{equation}
\mathcal{S} = \{  \pos \in \Real^{3} \vert g(\pos) = 0 \}.
\end{equation}

The SDF is parameterized by a fully connected neural network $g(\encOut_g; \NetParams_g)$ that processes an encoding ${\encOut_g = \enc(\pos; \LatticeParams_g)}$ of the input position $\pos$.  We refer to the composition $g(\enc(\pos; \LatticeParams_g); \NetParams_g)$ as the SDF network which outputs SDF values for a given spatial 3D position $\pos$.

Similarly, we define an MLP $c(\encOut_c, \ViewDir, \normal, \GeomFeat; \NetParams_c)$ for the color which processes an encoded position  ${\encOut_c = \enc(\pos; \LatticeParams_c)}$, a view direction $\ViewDir$, the normal vector of the SDF $\normal$,  and a learnable geometric feature $\GeomFeat$ which is output by the SDF network.

\reffig{fig:overview} given an overview of our two-network pipeline. Rays from the input images are cast into the scenes and multiple samples are created along each ray. Each sample $\mathbf{x}$ on the ray is encoded using a multi-resolutional hash-based permutohedral lattice (cf.~\refsec{sec:hashencode}).
The lattice features from different levels are concatenated and processed by the color and SDF MLPs.  
The SDF values are mapped to density (cf.~\refsec{sec:volRend}) and the sample colors are rendered volumetrically to yield the final pixel value.

Note, that using two separate networks is crucial as we want to regularize each one individually in order to recover high-quality geometry.

\section{Permutohedral Lattice SDF Rendering}

We now detail each network, the permutohedral lattice, and our training methodology.

\subsection{Volumetric Rendering}

\label{sec:volRend}

We denote the ray emitted from a pixel by  
$ \Pixel(t)= \CamOrig + t\ViewDir$, where $\CamOrig$ is the camera origin and $\ViewDir$ is the view direction. Colors along the ray are accumulated according to 
\begin{equation}
\label{eq:colorInt}
	\hat{C}(\Pixel) = \int_{t=0}^{+\infty} w(t) c(\Pixel(t), \ViewDir, \normal, \GeomFeat; \NetParams_c),
\end{equation}
where $w(t)$ is a weighting function for the point at $\Pixel(t)$.

In NeuS~\cite{neus}, Wang et al. show that in order to learn an SDF of the scene, it is crucial to derive an appropriate weighting function based on the SDF.

They propose an unbiased and occlusion-aware weighting function based on an opaque density function $\rho(t)$:

\begin{equation}
\begin{split}
\rho(t) =& \max\left(\frac{-\frac{{\rm d}\psi_s}{{\rm d} t}( g(\mathbf{p}(t);\NetParams_g)   )}{\psi_s( g(\mathbf{p}(t);\NetParams_g) )}, 0\right),
\end{split}
\label{eq:sigma}
\end{equation}
where $g(\mathbf{p}(t))$ outputs the SDF for the point at $t$ and $\psi$ is the sigmoid function defined as $\psi_s(x) = (1 + e^{-\SigmoidSlope x})^{-1}$ with slope $\SigmoidSlope$.
This can be used directly in a volumetric rendering scheme:
\begin{equation}
\!w(t) = T(t)\rho(t), \; {\rm where } \  T(t) =  \exp\left(-\!\int_{0}^{t}\!\rho(u){\rm d}u\right)
\label{eq:new_weight}
\end{equation}

\subsection{Hash Encoding with Permutohedral Lattice}
\label{sec:hashencode}
In order to facilitate learning of high-frequency details, INGP~\cite{ingp} proposed a hash-based encoding which maps a 3D spatial coordinate to a higher-dimensional space. The encoding maps a spatial position $\pos$ into a cubical grid and linearly interpolates features from the hashed eight corners of the containing cube. 
A fast CUDA implementation interpolates over various multi-resolutional grids in parallel. 
The hash map is stored as a tensor of $\levels$ levels, each containing up to $\entriesPerLevel$ feature vectors with dimensionality $\featuresPerEntry$. 

The speed of the encoding function is mostly determined by the number of accesses to the hash map as the operations to determine the eight hash indices are fast. Hence, it is of interest to reduce the memory accesses required to linearly interpolate features for position $\pos$. By using a tetrahedral lattice instead of a cubical one, memory accesses can be reduced by a factor of two as each simplex has only four vertices instead of eight. This advantage grows for higher dimensions when using a permutohedral lattice~\cite{adams2010fast}. 

The permutohedral lattice divides the space into uniform simplices which form triangles and tetrahedra in 2D and 3D, respectively. The main advantage of this lattice is that given dimensionality $d$ the number of vertices per simplex is $d+1$, which scales linearly instead of the exponential growth $2^d$ for hyper-cubical voxels. This ensures a low number of memory accesses to the hashmap and therefore fast optimization.

Given a position $\pos$, the containing simplex can be obtained in $\BigO(d^2)$. Within the simplex, barycentric coordinates are calculated and d-linear interpolation is performed similar to INGP. For more details regarding calculating the containing simplex, we refer to~\cite{permutoproperties}.

Similarly to INGP, we slice from lattices at multiple resolutions and concatenate the results. The final output is a high-dimensional encoding ${\encOut = \enc(\pos; \LatticeParams)}$ of the input $\pos$ given lattice features $\LatticeParams$.

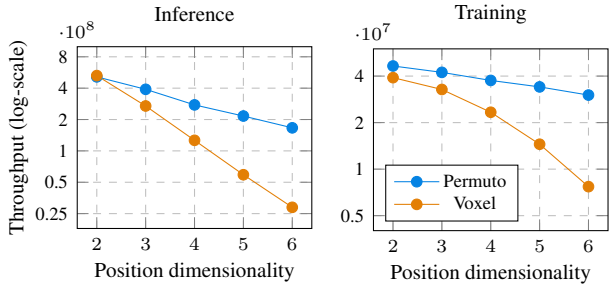
\begin{figure}

\begin{minipage}{.5\linewidth}
	\centering
	\begin{tikzpicture} [trim left=-1.2cm]
	
	\begin{axis} [
	bar width = 6pt,
	height=4cm,
	width=4.7cm,
	ymode=log,
	log basis y={2},
  ticklabel style = {font=\scriptsize},
	ylabel = {\footnotesize Throughput (log-scale) },
	ylabel shift = -4pt,
	xlabel shift = -2pt,
	xlabel = {\footnotesize Position dimensionality},
	grid=major,grid style={dashed},
	xtick={2,3,4,5,6},
	ymax=1000000000,
	ymin=18000000,
	ytick={800000000,400000000,200000000,100000000, 50000000,25000000},
	yticklabels={8,4,2,1,0.5, 0.25},
	title={\footnotesize Inference},
	legend style={nodes={scale=0.7, transform shape}}
	]

	\addplot [ph-blue, mark=*] coordinates {

		(2, 513506650)
		(3, 389956566)
		(4, 276393834)
		(5, 216044507)
		(6, 166469158)

	};

	\addplot [ph-orange, mark=*] coordinates {

		(2, 528055093)
		(3, 269811292)
		(4, 126301502)
		(5, 59039792)
		(6, 28730444)
		
	};

	\end{axis}

	\node at (0, 2.65) {\scriptsize\(\cdot10^{8}\)}; %
	
	\end{tikzpicture}
\end{minipage}%
\begin{minipage}{.5\linewidth}
	\centering
	\begin{tikzpicture}  [trim left=-0.9cm]
	
	\begin{axis} [
	bar width = 6pt,
	height=4cm,
	width=4.7cm,
	ymode=log,
	log basis y={2},
	ylabel shift = -2pt,
	xlabel shift = -2pt,
	xlabel = {\footnotesize Position dimensionality},
	grid=major,grid style={dashed},
	xtick={2,3,4,5,6},
	ymax=60000000,
	ymin=4000000,
	ticklabel style = {font=\scriptsize},
  ytick={40000000,20000000,10000000,5000000},
	yticklabels={4,2,1,0.5},
	title={\footnotesize Training},
	legend style={
		nodes={scale=0.7, transform shape},
		at={(0.05,0.05)},anchor=south west
		}
	]

	\addplot [ph-blue, mark=*] coordinates {

		(2, 46494717)
		(3, 42278864)
		(4, 37493257)
		(5, 34047975)
		(6, 30144309)
	};
	\addplot [ph-orange, mark=*] coordinates {

		(2, 39084521)
		(3, 32767769)
		(4, 23336648)
		(5, 14493888)
		(6, 7705060)
	};

	\addlegendentry{Permuto}
	\addlegendentry{Voxel}

	\end{axis}

	\node at (0, 2.65) {\scriptsize\(\cdot10^{7}\)}; %
	
	\end{tikzpicture}
\end{minipage} 
\vspace{-3mm}
\caption{ 
We use a permutohedral lattice instead of hyper-cubical voxels since the number of vertices per simplex scales linearly with the dimensionality instead of exponentially.  
The permutohedral lattice trains faster and encodes points faster for dim. $\ge 3$.
}
\label{fig:performance_lattice}
\vspace{-4mm}
\end{figure}

\subsection{4D Background Estimation}

For modeling the background, we follow the formulation of NeRF++~\cite{nerf++} which represents foreground volume as a unit sphere and background volume by an inverted sphere. Points in the outer volume are represented using 4D positions $(x',y',z', 1/r)$ where $(x',y',z')$ is a unit-length directional vector and $1/r$ is the inverse distance. 

We directly use this 4D coordinate to slice from a 4-dimensional lattice and obtain multi-resolutional features. A small MLP outputs the radiance and density which are volumetrically rendered and blended with the foreground. Please note that in 4D, the permutohedral lattice only needs to access five vertices for each simplex while a cubical voxel would need 16. Our linear scaling with dimensionality is of significant advantage in this use case.

\section{PermutoSDF Training and Regularization}
	Given the permutohedral lattice hash encoding and the unbiased volumetric rendering scheme, we have all the tools to train our model. 
	We sample pixels from the input images and infer SDF and color for positions along their rays. Through volumetric rendering (Eq.~\ref{eq:colorInt}) we obtain the pixel color $\hat{C}(\Pixel)$. 	
	We optimize an L2 loss on the RGB pixels:
	\begin{equation}
	\Loss_\textrm{rgb}= \sum_{p} \norm{ \hat{C}(\Pixel) - C(\Pixel)  }^2_2
	\end{equation}
	and an Eikonal loss which prevents the zero-everywhere solution for the SDF:
	\begin{equation}
	\Loss_\textrm{eik}= \sum_{x}  \left(  \norm{  \nabla g(\enc(\pos))    }  -1   \right)^2, 
	\end{equation}
	
	where the gradient $\nabla g(\enc(\pos))$ of the SDF is obtained through automatic differentiation.

	A naive combination of hash-based encoding and implicit surfaces can yield undesirable surfaces, though. While the model is regularized by the Eikonal loss, there are many surfaces that satisfy the Eikonal constraint. For specular or untextured areas, the Eikonal regularization doesn't provide enough information to properly recover the surface. 
	To address this issue, we propose several regularizations that serve to both recover smoother surfaces and more detail. 

\subsection{SDF Regularization}

In order to aid the network in recovering smoother surfaces in reflective or untextured areas, we add a curvature loss on the SDF. Calculating the full 3$\times$3 Hessian matrix can be expensive; so we approximate curvature as local deviation of the normal vector. Recall that we already have the normal $\normal =  \nabla g(\enc(\pos))$ at each ray sample since it was required for the Eikonal loss. With this normal, we define a tangent vector $\tangent$ by cross product with a random unit vector $\RandomVec$ such that
$\tangent = \normal \times \RandomVec$.
Given this random vector in the tangent plane, we slightly perturb our sample $\pos$ to obtain $\pos_{\epsilon} = \pos + \epsilon \tangent$.
We obtain the normal at the new perturbed point as $\normal_{\epsilon} =  \nabla g(\enc(\pos_{\epsilon}))$ and define a curvature loss based on the dot product between the normals at the original and perturbed points: 

\begin{align}
\Loss_\textrm{curv} &= \sum_{x} ( \normal \cdot \normal_{\epsilon}  -1) ^2.
\end{align}

\subsection{Color Regularization}

While the curvature regularization helps in recovering smooth surfaces, we observe that the network converges to an undesirable state where the geometry gets increasingly smoother while the color network learns to model all the high-frequency detail in order to drive the $\Loss_\textrm{rgb}$ to zero. Despite lowering $\Loss_\textrm{curv}$ during optimization, the SDF doesn't regain back the lost detail. We show this behavior in~\reffig{fig:ablation_study}. Recall that the color network is defined as  $c(\encOut, \ViewDir, \normal, \GeomFeat; \NetParams_c)$,  with an input encoding of ${\encOut = \enc(\pos; \LatticeParams_c)}$. We observe that all the high-frequency detail learned by the  color network has to be present in the  weights of the MLP $\NetParams_c$ or the hash-map table $\LatticeParams_c$ as all the other inputs are smooth.

In order to recover fine geometric detail, we propose to learn a color mapping network that is itself smooth \wrt to its input such that large changes in color are matched with large changes in surfaces normal. Function smoothness can be studied in the context of Lipschitz continuous networks. A function $f$ is k-Lipschitz continuous if it satisfies:
\begin{align}
\underbrace{\| f(d) - f(e) \|}_{\mathclap{\text{change in the output}}} \leq \lipc\ \underbrace{\| d - e\|}_{\mathclap{\text{change in the input}}}.
\end{align}
Intuitively, it sets $k$ as an upper bound for the rate of change of the function. 
We are interested in the color network being a smooth function (small $k$) such that high-frequency color is also reflected in high-detailed geometry.

There are several ways to enforce Lipschitz smoothness on a function~\cite{miyato2018spectral,cisse2017parseval,terjek2019adversarial,yoshida2017spectral}. Most of them impose a hard 1-Lipschitz requirement or ignore effects such as network depth which makes them difficult to tune for our use case.

The recent work of Liu \etal~\cite{liu2022learning} provides a simple and interpretable framework for softly regularizing the Lipschitz constant of a network. 
Given an MLP layer $y=\sigma(W_ix+b_i)$ and a trainable Lipschitz bound $k_i$ for the layer, they replace the weight matrix $W_i$ with:
\begin{align}
y = \sigma( \widehat{W}_i x + b_i), \quad \widehat{W}_i = \textit{m}\,(W_i, \softplus(k_i)),
\end{align}
where $\softplus(k_i) = \ln(1+e^{k_i})$ and the function $\textit{m}(.)$ normalizes the weight matrix by rescaling each row of $W_i$ such that the absolute value of the row-sum is less than or equal to $\softplus(k_i)$.
Since the product of per-layer Lipschitz constants $k_i$ is the Lipschitz bound for the whole network, we can regularize it using:
\begin{align}
\Loss_\textrm{Lipschitz}= \prod_{l}^{i=1} \softplus(k_i).
\end{align}

In addition to regularizing the color MLP, we also apply weight decay of $0.01$ to the color hashmap $\LatticeParams_c$.

\begin{figure}
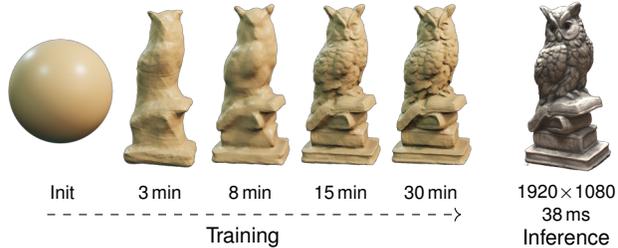


\small\sffamily
\centering
\begin{tikzpicture}

\newcommand\ShiftXLarge{1.2} %
\newcommand\ShiftYLarge{-1.4}

\node[inner sep=0pt] (d0) at (\ShiftXLarge*0, 0)
    {\includegraphics[
    	width=.18\columnwidth
    	]{./imgs/progress/cropped/sphere2.png}};
\node[inner sep=0pt] (d0) at (\ShiftXLarge*1,0)
{\includegraphics[
	width=.25\columnwidth
	]{./imgs/progress/cropped/mesh_3min.png}};
\node[inner sep=0pt] (d0) at (\ShiftXLarge*2,0)
{\includegraphics[
	width=.25\columnwidth
	]{./imgs/progress/cropped/mesh_8min.png}};
\node[inner sep=0pt] (d0) at (\ShiftXLarge*3,0)
{\includegraphics[
	width=.25\columnwidth
	]{./imgs/progress/cropped/mesh_15min.png}};
\node[inner sep=0pt] (d0) at (\ShiftXLarge*4,0)
{\includegraphics[
	width=.25\columnwidth
	]{./imgs/progress/cropped/mesh_30min.png}};

\node[inner sep=0pt] (d0) at (\ShiftXLarge*5+ 0.7, 0)
{\includegraphics[
	width=.16\columnwidth
	]{./imgs/progress/cropped/img_sphere_trace_enhanced2.png}};

\node[] at (\ShiftXLarge*0, \ShiftYLarge) {\scriptsize Init};
\node[] at (\ShiftXLarge*1+0.1, \ShiftYLarge) {\scriptsize {\SI{3}{\minute}}};
\node[] at (\ShiftXLarge*2+0.1, \ShiftYLarge) {\scriptsize {\SI{8}{\minute}}};
\node[] at (\ShiftXLarge*3+0.1, \ShiftYLarge) {\scriptsize {\SI{15}{\minute}}};
\node[] at (\ShiftXLarge*4+0.1, \ShiftYLarge) {\scriptsize {\SI{30}{\minute}}};
\node[] at (\ShiftXLarge*5+0.7, \ShiftYLarge) {\scriptsize \num{1920}$\times$\num{1080} };
\node[] at (\ShiftXLarge*5+0.7, \ShiftYLarge-0.3) {\scriptsize {\SI{38}{\milli\second}} };
\node[] at (\ShiftXLarge*2, \ShiftYLarge-.6) {\footnotesize Training};
\node[] at (\ShiftXLarge*5+0.7, \ShiftYLarge-.6) {\footnotesize Inference};

\draw[->, dashed] (0-0.2, \ShiftYLarge-0.3) -- (5+0.3,\ShiftYLarge-0.3) node[above] {};

\end{tikzpicture}
\vspace{-3.5mm}
\caption{ We train for \SI{30}{\minute} using posed images and afterwards render novel views in real-time using sphere tracing.  } \label{fig:progress}
\vspace{-2mm}
\end{figure} 

\subsection{Training Schedule}
Several scheduling considerations must be observed for our method. 
In~\refequ{eq:sigma}, the sigmoid function $\psi_s(.)$ is parametrized with  $1/\SigmoidSlope$ which is the standard deviation that controls the range of influence of the SDF towards the volume rendering. 
In NeuS $1/\SigmoidSlope$ is considered a learnable parameter which starts at a high value and decays towards zero as the network converges.

However, we found out that considering it as a learnable parameter can lead to the network missing thin object features due to the fact that large objects in the scene dominate the gradient towards $\SigmoidSlope$. 
Instead, we use a scheduled linear decay $1/\SigmoidSlope$ over \SI{30}{\kilo{}} iterations which we found to be robust for all the objects we tested.

In order to recover smooth surfaces, we train the first \SI{100}{\kilo{}} iterations using curvature loss:
\begin{equation}
\Loss= \Loss_\textrm{rgb} + \lambda_{1} \Loss_\textrm{eik} + \lambda_2 \Loss_\textrm{curv}.
\end{equation} 
For further \SI{100}{\kilo{}} iterations, we recover detail by removing the curvature loss and adding the regularization of the color network $\lambda_3 \Loss_\textrm{Lipschitz}$.

In addition, we initialize our network with the SDF of a sphere at the beginning of the optimization and anneal the levels $\levels$ of the hash map in a coarse-to-fine manner over the course of the initial \SI{10}{\kilo{}} iterations.
We refer to the supplementary material for more details regarding the hyperparameters $\lambda_{1-3}$.

\section{Acceleration}	
Similar to other volumetric rendering methods, a major bottleneck for the speed is the number of position samples considered for each ray. We use several methods to accelerate both training and inference.

\begin{figure*}[h!]
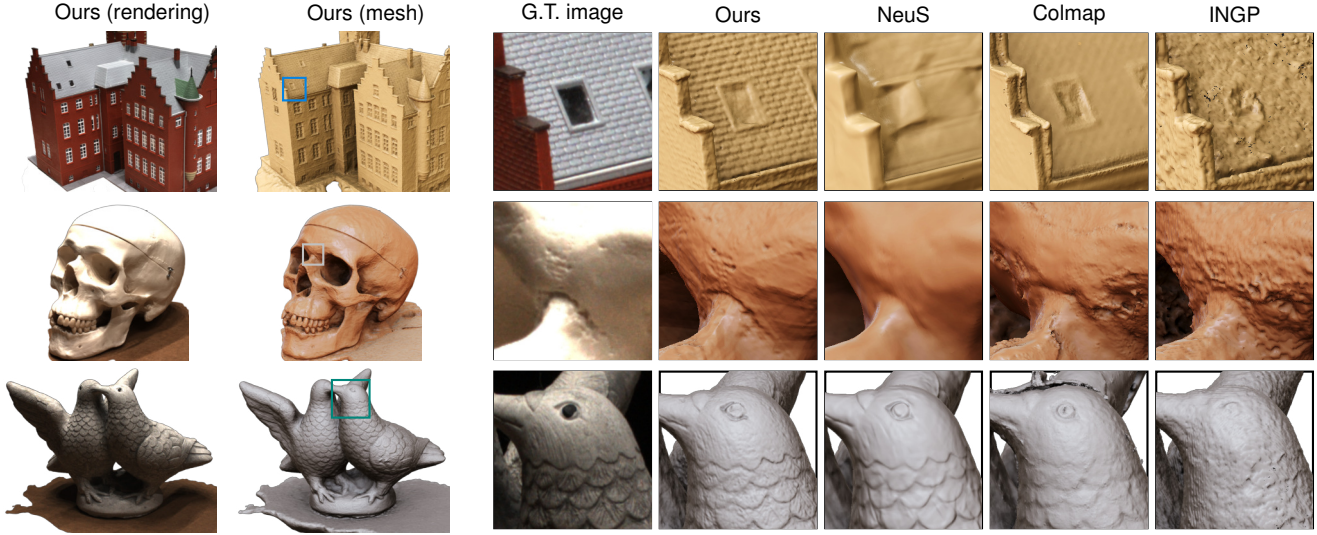

\centering
\small\sffamily
\begin{tikzpicture}[square/.style={regular polygon,regular polygon sides=4}]

\newcommand\ShiftXLarge{3.1} %
\newcommand\ShiftXSmall{2.2} %
\newcommand\ShiftXZomminStart{6.15} %
\newcommand\ShiftySmall{0.0} %
\newcommand\ShiftyLarge{-2.25} %

\node[inner sep=0pt] (d0) at (\ShiftXLarge*0,0)
    {\includegraphics[
    	width=.34\columnwidth
    	]{./imgs/dtu_mesh_comparison/dtu_scan24/34rgb_alpha.png}};
\node[inner sep=0pt] (d0) at (\ShiftXLarge*1,0)
{\includegraphics[
	width=.34\columnwidth
	]{./imgs/dtu_mesh_comparison/dtu_scan24/mine/rendered_fullTrue_subsample1.png}};

\node[inner sep=0pt, draw, thick] (d0) at (\ShiftXZomminStart+\ShiftXSmall*0, \ShiftySmall)
{\includegraphics[
	width=.25\columnwidth
	]{./imgs/dtu_mesh_comparison/dtu_scan24/zoomins/gt.png}};

\node[inner sep=0pt, draw, thick] (d0) at (\ShiftXZomminStart+\ShiftXSmall*1, \ShiftySmall)
{\includegraphics[
	width=.25\columnwidth
	]{./imgs/dtu_mesh_comparison/dtu_scan24/zoomins/mine.png}};
\node[inner sep=0pt, draw, thick] (d0) at (\ShiftXZomminStart+\ShiftXSmall*2, \ShiftySmall)
{\includegraphics[
	width=.25\columnwidth
	]{./imgs/dtu_mesh_comparison/dtu_scan24/zoomins/neus.png}};
\node[inner sep=0pt, draw, thick] (d0) at (\ShiftXZomminStart+\ShiftXSmall*3, \ShiftySmall)
{\includegraphics[
	width=.25\columnwidth
	]{./imgs/dtu_mesh_comparison/dtu_scan24/zoomins/colmap.png}};
\node[inner sep=0pt, draw, thick] (d0) at (\ShiftXZomminStart+\ShiftXSmall*4, \ShiftySmall)
{\includegraphics[
	width=.25\columnwidth
	]{./imgs/dtu_mesh_comparison/dtu_scan24/zoomins/ingp.png}};
\node[inner sep=0pt] (d0) at (\ShiftXLarge*0, \ShiftyLarge*1)
{\includegraphics[
	width=.25\columnwidth
	]{./imgs/dtu_mesh_comparison/dtu_scan65/cropped/mine_rgb_origRes.png}};
\node[inner sep=0pt] (d0) at (\ShiftXLarge*1, \ShiftyLarge*1)
{\includegraphics[
	width=.25\columnwidth
	]{./imgs/dtu_mesh_comparison/dtu_scan65/cropped/mine_origRes.png}};
\node[inner sep=0pt, draw, thick] (d0) at (\ShiftXZomminStart+\ShiftXSmall*0, \ShiftySmall+\ShiftyLarge*1)
{\includegraphics[
	width=.25\columnwidth
	]{./imgs/dtu_mesh_comparison/dtu_scan65/zoomins/3/gt.png}};
\node[inner sep=0pt, draw, thick] (d0) at (\ShiftXZomminStart+\ShiftXSmall*1, \ShiftySmall+\ShiftyLarge*1)
{\includegraphics[
	width=.25\columnwidth
	]{./imgs/dtu_mesh_comparison/dtu_scan65/zoomins/3/mine.png}};
\node[inner sep=0pt, draw, thick] (d0) at (\ShiftXZomminStart+\ShiftXSmall*2, \ShiftySmall+\ShiftyLarge*1)
{\includegraphics[
	width=.25\columnwidth
	]{./imgs/dtu_mesh_comparison/dtu_scan65/zoomins/3/neus.png}};
\node[inner sep=0pt, draw, thick] (d0) at (\ShiftXZomminStart+\ShiftXSmall*3, \ShiftySmall+\ShiftyLarge*1)
{\includegraphics[
	width=.25\columnwidth
	]{./imgs/dtu_mesh_comparison/dtu_scan65/zoomins/3/colmap.png}};
\node[inner sep=0pt, draw, thick] (d0) at (\ShiftXZomminStart+\ShiftXSmall*4, \ShiftySmall+\ShiftyLarge*1)
{\includegraphics[
	width=.25\columnwidth
	]{./imgs/dtu_mesh_comparison/dtu_scan65/zoomins/3/ingp.png}};

\node[inner sep=0pt] (d0) at (\ShiftXLarge*0, \ShiftyLarge*2)
{\includegraphics[
	width=.34\columnwidth
	]{./imgs/dtu_mesh_comparison/dtu_scan106/cropped/mineRgb_cropped_bright_origRes.png}};
\node[inner sep=0pt] (d0) at (\ShiftXLarge*1, \ShiftyLarge*2)
{\includegraphics[
	width=.34\columnwidth
	]{./imgs/dtu_mesh_comparison/dtu_scan106/cropped/mine_cropped_origRes.png}};

\node[inner sep=0pt, draw, thick] (d0) at (\ShiftXZomminStart+\ShiftXSmall*0, \ShiftySmall+\ShiftyLarge*2)
{\includegraphics[
	width=.25\columnwidth
	]{./imgs/dtu_mesh_comparison/dtu_scan106/zoomins/gt.png}};
\node[inner sep=0pt, draw, thick] (d0) at (\ShiftXZomminStart+\ShiftXSmall*1, \ShiftySmall+\ShiftyLarge*2)
{\includegraphics[
	width=.25\columnwidth
	]{./imgs/dtu_mesh_comparison/dtu_scan106/zoomins/mine.png}};
\node[inner sep=0pt, draw, thick] (d0) at (\ShiftXZomminStart+\ShiftXSmall*2, \ShiftySmall+\ShiftyLarge*2)
{\includegraphics[
	width=.25\columnwidth
	]{./imgs/dtu_mesh_comparison/dtu_scan106/zoomins/neus.png}};
\node[inner sep=0pt, draw, thick] (d0) at (\ShiftXZomminStart+\ShiftXSmall*3, \ShiftySmall+\ShiftyLarge*2)
{\includegraphics[
	width=.25\columnwidth
	]{./imgs/dtu_mesh_comparison/dtu_scan106/zoomins/colmap.png}};
\node[inner sep=0pt, draw, thick] (d0) at (\ShiftXZomminStart+\ShiftXSmall*4, \ShiftySmall+\ShiftyLarge*2)
{\includegraphics[
	width=.25\columnwidth
	]{./imgs/dtu_mesh_comparison/dtu_scan106/zoomins/ingp.png}};

\node[] at (\ShiftXLarge*0+.3,1.3) {\footnotesize Ours (rendering) };
\node[] at (\ShiftXLarge*1+.3,1.3) {\footnotesize Ours (mesh) };
\node[] at (\ShiftXZomminStart+\ShiftXSmall*0,1.3) {\footnotesize G.T. image };
\node[] at (\ShiftXZomminStart+\ShiftXSmall*1,1.3) {\footnotesize Ours };
\node[] at (\ShiftXZomminStart+\ShiftXSmall*2,1.3) {\footnotesize NeuS };
\node[] at (\ShiftXZomminStart+\ShiftXSmall*3,1.3) {\footnotesize Colmap };
\node[] at (\ShiftXZomminStart+\ShiftXSmall*4,1.3) {\footnotesize INGP };

\node[rectangle,draw,ph-blue, thick, minimum width = 0.3cm, minimum height = 0.3cm] (r) at (2.45,0.3) {};
\node[rectangle,draw,ph-silver, thick, minimum width = 0.27cm, minimum height = 0.27cm] (r) at (2.7,-1.9) {};
\node[rectangle,draw,ph-green, thick, minimum width = 0.5cm, minimum height = 0.5cm] (r) at (3.2,-3.82) {};

\end{tikzpicture}
\vspace{-5mm}
\caption{ Qualitative comparison of the geometry reconstructed by our method compared to the baselines. Note that our method recovers significantly higher geometrical detail.  } \label{fig:dtu_mesh_comparison}
\vspace{-3mm}
\end{figure*} 
\subsection{Occupancy Grid} 
In order to concentrate more samples near the surface of the SDF and have fewer in empty space, we use an occupancy grid modeled as a dense grid of resolution $128^3$.
 
We maintain two versions of the grid, one with full precision, storing the SDF value at each voxel, and another containing only a binary occupancy bit. The grids are laid out in Morton order to ensure fast traversal. Note that differently from INGP, we store signed distance and not density in our grid. This allows us to use the SDF volume rendering equations to determine if a certain voxel has enough weight that it would meaningfully contribute to the integral~\refequ{eq:colorInt} and therefore should be marked as occupied space.

We refer to the supplementary material for more details on the update of the occupancy grid.

\subsection{Sphere Tracing}
Having an SDF opens up a possibility for accelerating rendering at inference time by using sphere tracing. This can be significantly faster than volume rendering as most rays converge in 2-3 iterations towards the surface. We create a ray sample at the first voxel along the ray that is marked as occupied. We run sphere tracing for a predefined number of iterations and march not-yet-converged samples towards the surface (indicated by their SDF being above a certain threshold). Once all samples have converged or we reached a maximum number of sphere traces, we sample the color network once and render. 

We show in~\reffig{fig:progress} that by using sphere tracing we can render in real time and can also trade-off rendering speed against accuracy by varying the number of iterations.

\subsection{Implementation Details} 

We implement the encoding ${\encOut = \enc(\pos; \LatticeParams)}$ using permutohedral lattices in a custom CUDA kernel which slices in parallel from all resolutions. The backward pass for updating the hashmap $\frac{\partial \encOut}{ \partial \LatticeParams }$ is also implemented in an optimized CUDA kernel. We use the chain rule to backpropagate the upstream gradients as: $ \frac{\partial\Loss}{\partial \encOut}   \frac{\partial \encOut}{ \partial \LatticeParams }$.

Additionally, since we require the normals ${\normal=\nabla g(\enc(\pos))} $ for fitting the SDF, we also implement a kernel for calculating the partial derivative of encoding ${\encOut = \enc(\pos; \LatticeParams_g)}$ \wrt to spatial position $\pos$, i.e. $\frac{\partial \encOut}{ \partial \pos }$. Again, the chain rule is applied with the autograd partial derivative of $g(.)$ as: $ \frac{\partial g}{\partial \encOut}   \frac{\partial \encOut}{ \partial \pos }$
to obtain the normal.

Furthermore, since we use this normal as part of our loss function $\Loss_\textrm{eik}$, we support also double backward operations, i.e., we also implement CUDA kernels for $ \sfrac{ \partial(   \frac{\partial \Loss}{\partial \pos}   )   }{\partial  \LatticeParams}  $ and 
$ \sfrac{ \partial(   \frac{\partial \Loss}{\partial \pos}   )   }{\partial  (\frac{\partial \Loss}{\partial \encOut})  }  $
Hence, we can run our optimization entirely within PyTorch's autograd engine, without requiring any finite differences.

\section{Results}

We evaluate our method on multiple datasets and report metrics on the accuracy of both, the 3D reconstruction and novel-view synthesis (NVS).

\subsection{DTU Data Set}
We evaluate the quality of 3D reconstruction on the DTU~\cite{dtu} dataset, which consists of 2D images of objects and ground-truth 3D point clouds. 
The objects span a wide range of materials with different specularities, which can pose a challenge for classical multi-view reconstruction methods.
Each object is captured with $\approx50$ images and we use every 8-th image for testing and the rest for training. We evaluate against NeuS~\cite{neus} which is our baseline, INGP~\cite{ingp} which is state-of-the-art in NVS rendering, and COLMAP~\cite{colmap}, a classical multi-view stereo method. 

We show qualitative results of the extracted meshes in~\reffig{fig:dtu_mesh_comparison}, where we train all methods without any mask supervision. Our method surpasses the level-of-detail of the other methods and is robust to view-dependent and untextured areas. 

We report quantitative Chamfer distance results in~\reftab{tbl:dtu_chamfer}, comparing reconstruction with and without mask supervision.
In both cases, our method significantly outperforms the competing approaches.

\begin{figure}
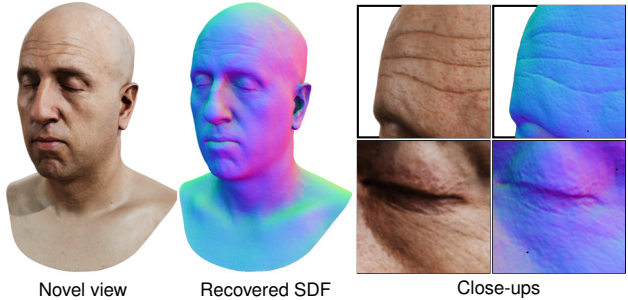

\centering
\small\sffamily
\begin{tikzpicture}

\newcommand\ShiftXLarge{2.3} %
\newcommand\ShiftXZomminStart{4.4} %
\newcommand\ShiftXSmall{1.8}
\newcommand\Shifty{.9} 

\node[inner sep=0pt] (d0) at (\ShiftXLarge*0,0)
    {\includegraphics[
    	width=.28\columnwidth
    	]{./imgs/head_synth/pred_img_matTrue3_cw.png}};
\node[inner sep=0pt] (d0) at (\ShiftXLarge*1,0)
{\includegraphics[
	width=.28\columnwidth
	]{./imgs/head_synth/rendered_fullTrue3_cw.png}};
\node[inner sep=0pt, draw,  thick] (d0) at (\ShiftXZomminStart+\ShiftXSmall*0,\Shifty*1)
{\includegraphics[
	width=.21\columnwidth
	]{./imgs/head_synth/pred_img_matTrue4.png}};
\node[inner sep=0pt, draw,  thick] (d0) at (\ShiftXZomminStart+\ShiftXSmall*1,\Shifty*1)
{\includegraphics[
	width=.21\columnwidth
	]{./imgs/head_synth/rendered_fullTrue4.png}};

\node[inner sep=0pt, draw, thick] (d0) at (\ShiftXZomminStart+\ShiftXSmall*0,-\Shifty*1)
{\includegraphics[
	width=.21\columnwidth
	]{./imgs/head_synth/pred_img_matTrue1.png}};
\node[inner sep=0pt, draw, thick] (d0) at (\ShiftXZomminStart+\ShiftXSmall*1,-\Shifty*1)
{\includegraphics[
	width=.21\columnwidth
	]{./imgs/head_synth/rendered_fullTrue2.png}};

\node[] at (\ShiftXLarge*0,-2.) {\scriptsize \hspace*{-2ex}Novel view };
\node[] at (\ShiftXLarge*1,-2.) {\scriptsize Recovered SDF };
\node[] at (\ShiftXLarge*2+0.8,-2.) {\scriptsize Close-ups };

\end{tikzpicture}
\vspace{-7mm}
\caption{ Given high-resolution synthetically rendered images, our approach can recover small details like pores and wrinkles. Please refer to the suppl. material for reconstructions of other methods. } \label{fig:head_synth}
\vspace*{-2mm}
\end{figure} 
\begin{table}
	\centering
	\resizebox{\linewidth}{!}{
		\begin{threeparttable}
			\begin{tabular}{c||c|c|c||c|c|c|c}
				\multicolumn{1}{c||}{}&\multicolumn{3}{c||}{\textbf{w/} mask}&\multicolumn{4}{c}{\textbf{w/o} mask} \\
				\hline
				 & INGP & NeuS & Ours & \hspace*{-1.5ex}  COLMAP \hspace*{-1.5ex} & INGP & NeuS & Ours \vspace*{-0.5ex}\\
				ScanID & \cite{ingp} & \cite{neus} &  &   \cite{colmap} & \cite{ingp} & \cite{neus} &  \\
				\hline
				scan24 & 1.73 & 0.83 & \bf{0.53} &    0.81 & 1.56 & 1.00 & \bf{0.52} \\
				scan37 & 1.79 & 0.98 & \bf{0.67} &      2.05 & 2.15 & 1.37 & \bf{0.75}  \\
				scan40 & 1.46 & 0.56 & \bf{0.34} &   0.73 & 1.45 & 0.93 & \bf{0.41}  \\
				scan55 & 0.86 & \bf{0.37} & 0.37 &    1.22 & 0.76 & 0.43 & \bf{0.37}  \\
				scan63 & 1.70 & 1.13 & \bf{0.94} &    1.79 & 1.62 & 1.10 & \bf{0.90}  \\
				scan65 & 1.57 & \bf{0.59} & 0.59 &    1.58 & 1.33 & \bf{0.65} & 0.66 \\
				scan69 & 1.66 & 0.60 & \bf{0.57} &    1.02 & 1.63 & \bf{0.57} & 0.59  \\
				scan83 & 1.56 & 1.45 & \bf{1.22} &    3.05 & 1.79 & 1.48 & \bf{1.37}  \\
				scan97 & 1.83 & 0.95 & \bf{0.78} &    1.40 & 2.16 & 1.09 & \bf{1.07}  \\
				scan105 & 1.55 & 0.78 & \bf{0.66} &    2.05 & 1.45 & \bf{0.83} & 0.85  \\
				scan106 & 1.23 & 0.52 & \bf{0.49} &    1.00 & 1.25 & 0.52 & \bf{0.46}  \\
				scan110 & 1.75 & 1.43 & \bf{0.73} &    1.32 & 1.91 & 1.20 & \bf{0.98}  \\
				scan114 & 1.71 & 0.36 & \bf{0.35} &    0.49 & 1.76 & 0.35 & \bf{0.33}  \\
				scan118 & 1.44 & 0.45 & \bf{0.41} &    0.78 & 1.24 & 0.49 & \bf{0.39}  \\
				scan122 & 1.31 & 0.49 & \bf{0.47} &    1.17 & 1.47 & 0.54 & \bf{0.50}  \\
				\hline
				mean & 1.54 & 0.77 & \bf{0.61} &       1.36 & 1.57 & 0.84 & \bf{0.68}  \\
				
			\end{tabular}\vspace*{-1.5ex}
			
		\end{threeparttable}
	}
	\caption{Quantitative Chamfer distance evaluation on the DTU dataset. COLMAP results are achieved by trim=0.  }
	\label{tbl:dtu_chamfer}
\end{table}

Additionally, we evaluate the quality of novel-view synthesis on the same dataset in~
\reftab{tbl:dtu_nvs_psnr}. One can observe that we surpass NeuS, NeRF, and in most cases also INGP. This is due to the fact that our method reconstructs the underlying geometry more faithfully, making it easier to generalize the rendering to novel views. 

\begin{table}
	\centering
	\resizebox{\linewidth}{!}{	
		\begin{threeparttable}
			\renewcommand{\arraystretch}{1.2}
			\setlength{\tabcolsep}{1mm} 
			\begin{tabular}{c||c|c|c|c c c||c|c|c|c}
				 & NeuS & NeRF & INGP & Ours & &  & NeuS & NeRF & INGP  & Ours \vspace*{-1ex} \\
				ScanID & \cite{neus} & \cite{nerf} & \cite{ingp} &  & & ScanID & \cite{neus} & \cite{nerf} & \cite{ingp}  &  \\
				\cline{0-4}
				\cline{7-11}

				scan24 & 26.76 & 27.54 & 28.77 & \bf{30.06} &   &  scan97 & 29.30 & \bf{30.46} & 29.43 & 30.45  \\
	
				scan37 & 25.84 & 26.54 & 26.34 & \bf{27.29} &   &  scan105 & 34.50 & 35.51 & 36.20 & \bf{36.85}  \\
				
				scan40 & 27.25 & 28.53 & 28.97 & \bf{30.43} &    &  scan106 & 34.12 & 34.86 & 35.05 & \bf{36.27}  \\
			
				scan55 & 28.09 & 30.39 & 31.20 & \bf{32.45} &   &  scan110 & 32.46 & 32.87 & 32.16 & \bf{34.52}  \\
				
				scan63 & 34.24 & 35.25 & \bf{36.72} & 36.32 &    &  scan114 & 30.01 & 30.82 & 31.04 & \bf{31.26}  \\
				
				scan65 & 33.83 & 33.42 & \bf{34.13} & 34.00 &    &  scan118 & 36.73 & 36.87 & 37.91 & \bf{38.70}  \\
				
				scan69 & 29.94 & 30.22 & 29.63 & \bf{30.49} &    &  scan122 & 37.89 & 37.77 & 38.64 & \bf{39.74}  \\
				
				\cline{7-11}
				scan83 & 39.02 & 40.12 & 40.29 & \bf{40.81} &   &  Mean  & 31.99  & 32.74 & 33.10  & \bf{33.97}  \\

			\end{tabular}
			\vspace*{-1.5ex} 
		\end{threeparttable}
	}
	\caption{Quantitative PSNR comparisons on the task of novel view synthesis without mask supervision.  }
	\label{tbl:dtu_nvs_psnr}
\end{table}

\subsection{Multiface Data Set}

Reconstructing human figures is especially difficult as they exhibit multiple view-dependent effects and fine details that need to be captured. To evaluate this, we use the Multiface datset~\cite{multiface}. It consists of human subjects that were captured with a dome of $\approx40$ high-resolution cameras while performing various facial expressions. 

\reffig{fig:multiface_mesh} shows a qualitative comparison between our reconstruction and NeuS. Our method is more detailed than NeuS, but it still struggles with very fine detail like hair where it usually learns to create hair-like streaks in the geometry in order to model eyebrows and beards.

\begingroup
\begin{figure*}
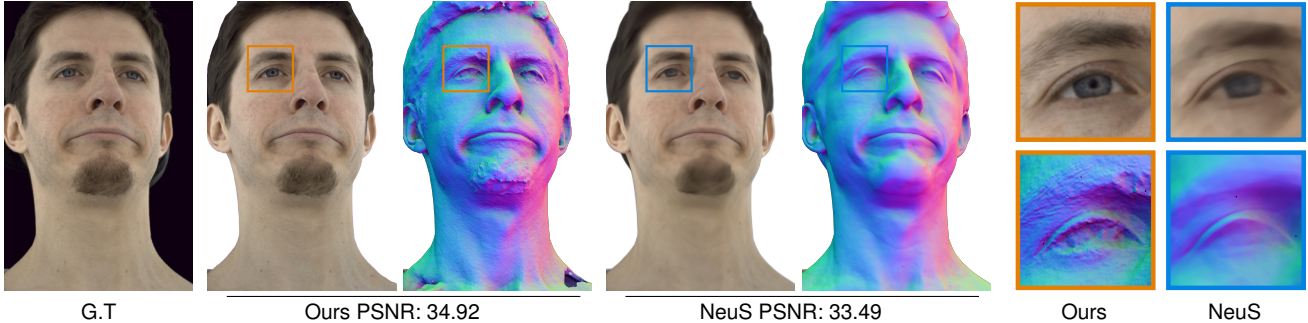

	\centering
	\small\sffamily
	\begin{tikzpicture}  [spy using outlines={ size=5cm,   every spy on node/.append style={thick}     }]

	\newcommand\ShiftXLarge{2.6} %
	\newcommand\ShiftyLarge{-2.5} %

	\node[inner sep=0pt] (gt) at (\ShiftXLarge*0,0)
	{\includegraphics[
		width=.3\columnwidth
		]{./imgs/multiface_mesh/5/gt_corr.png}};
	\node[inner sep=0pt] (minergb) at (\ShiftXLarge*1+.1,0)
	{\includegraphics[
		width=.3\columnwidth, compress=false
		]{./imgs/multiface_mesh/5/rgb_alpha_corr.png}};
	\node[inner sep=0pt] (minen) at (\ShiftXLarge*2+.1,0)
	{\includegraphics[
		width=.3\columnwidth, compress=false
		]{./imgs/multiface_mesh/5/5.png}};

	\node[inner sep=0pt] (neusrgb) at (\ShiftXLarge*3+.2,0)
	{\includegraphics[
		width=.3\columnwidth, compress=false
		]{./imgs/multiface_mesh/5/rgb_neus.png}};
	\node[inner sep=0pt] (neusn) at (\ShiftXLarge*4+.2,0)
	{\includegraphics[
		width=.3\columnwidth, compress=false
		]{./imgs/multiface_mesh/5/5_neus.png}};

	\spy [ph-orange,draw,height=1.8cm,width=1.8cm,magnification=3] on ($(minergb.center) + (-0.42, 1.03
	)$) in node [line width=0.6mm, anchor=north east] at ($(neusn.north east) + (-0.04+ 2.2, -0.04) $);
	\spy [ph-orange,draw,height=1.8cm,width=1.8cm,magnification=3] on ($(minen.center) + (-0.42, 1.03
	)$) in node [line width=0.6mm, anchor=south east] at ($(neusn.south east) + (-0.04+ 2.2, 0.04) $);

	\spy [ph-blue,draw,height=1.8cm,width=1.8cm,magnification=3] on ($(neusrgb.center) + (-0.42, 1.03
	)$) in node [line width=0.6mm, anchor=north east] at ($(neusn.north east) + (-0.04+ 4.2, -0.04) $);
	\spy [ph-blue,draw,height=1.8cm,width=1.8cm,magnification=3] on ($(neusn.center) + (-0.42, 1.03
	)$) in node [line width=0.6mm, anchor=south east] at ($(neusn.south east) + (-0.04+ 4.2, 0.04) $);

	\node[] at (\ShiftXLarge*0,-2.2) {\footnotesize G.T};
	\node[] at (\ShiftXLarge*1+.1+ 1.2,-2.2) {\footnotesize Ours PSNR: 34.92 };
	\node[] at (\ShiftXLarge*3+.2+ 1.2,-2.2) {\footnotesize NeuS PSNR: 33.49 };
	\node[] at (\ShiftXLarge*4+ 2.7,-2.2) {\footnotesize Ours};
	\node[] at (\ShiftXLarge*4+ 4.7,-2.2) {\footnotesize NeuS};

	\draw [] (\ShiftXLarge*1+.1-1.0,-2.0) -- (\ShiftXLarge*3+.1-1.5,-2.0);
	\draw [] (\ShiftXLarge*3+.2-1.0,-2.0) -- (\ShiftXLarge*5+.2-1.5,-2.0);

	\end{tikzpicture}
	\vspace{-2mm}
	\caption{ Qualitative comparison between our method and NeuS on the Multiface dataset. We can recover finer detail with higher PNSR.  } \label{fig:multiface_mesh}
\end{figure*}
\endgroup 
\subsection{Rendered Head Images}
Since the Multiface dataset was captured with real cameras, they exhibit several camera issues like depth-of-field effects and inaccurate color calibration which can prevent our method from learning more detail. To evaluate what can be achieved with perfect camera conditions, we render realistic images of a head figure~\cite{headsynth} using the EasyPBR renderer~\cite{easypbr}. Since the virtual cameras are perfectly calibrated and without defects, this can be seen as an upper bound on the quality that can be achieved with our method. We show in~\reffig{fig:head_synth} that we are capable of recovering details at the level of pores and wrinkles which cannot be achieved with previous learning-based volumetric rendering methods.

\subsection{Performance}
We evaluate the performance of our proposed permutohedral lattice for inference and training and compare it to the cubical voxels used in INGP. We encode a batch of $2^{19}$ random points and set both hash maps to a capacity $\entriesPerLevel\!=\!2^{18}$, $\levels\!=\!24$ levels, and $\featuresPerEntry\!=\!2$ features per entry.  In~\reffig{fig:performance_lattice} one can observe that our permutohedral lattice outperforms the cubical lattice during training and for dim.\,$>$2 during inference; with a larger gap for higher dimensions. This is to be expected since the number of vertices per simplex scales linearly with the dimensionality instead of exponentially. The only exception is in 2D, where a square lattice accesses four vertices per simplex while we access three---not much of an improvement and the cost of finding the simplex and calculating barycentric coordinates dominates.

\subsection{Ablation Study}

\begin{figure*}
\centering
\small\sffamily
\begin{tikzpicture} [spy using outlines={ size=5cm,   every spy on node/.append style={thick}     }]

\def\ABLImg{./imgs/ablation_study/ablation0.png} 
\newlength{\WABL}
\newlength{\HABL}
\settowidth{\WABL}{\includegraphics{\ABLImg}}
\settoheight{\HABL}{\includegraphics{\ABLImg}}

\newcommand\ShiftX{2.75}

\node[inner sep=0pt] (Abl0) at (0,0)
    {\includegraphics[
    	trim=.4\WABL{} .13\HABL{} .0\WABL{} .0\HABL{},clip,
    	width=.33\columnwidth, compress=false
    	]{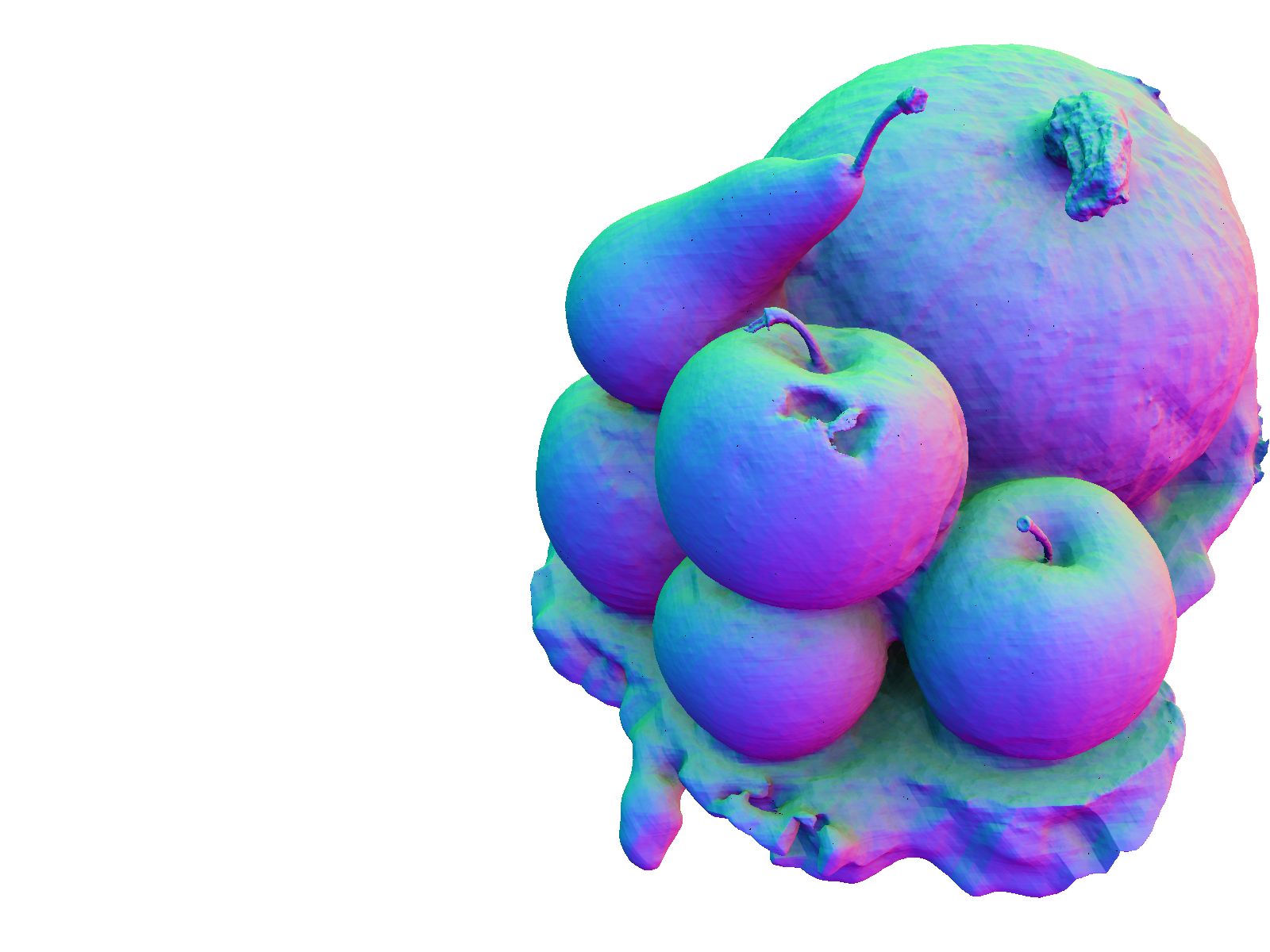}};
\node[inner sep=0pt] (Abl1) at (\ShiftX*1,0)
	{\includegraphics[
		trim=.4\WABL{} .13\HABL{} .0\WABL{} .0\HABL{},clip,
		width=.33\columnwidth, compress=false
		]{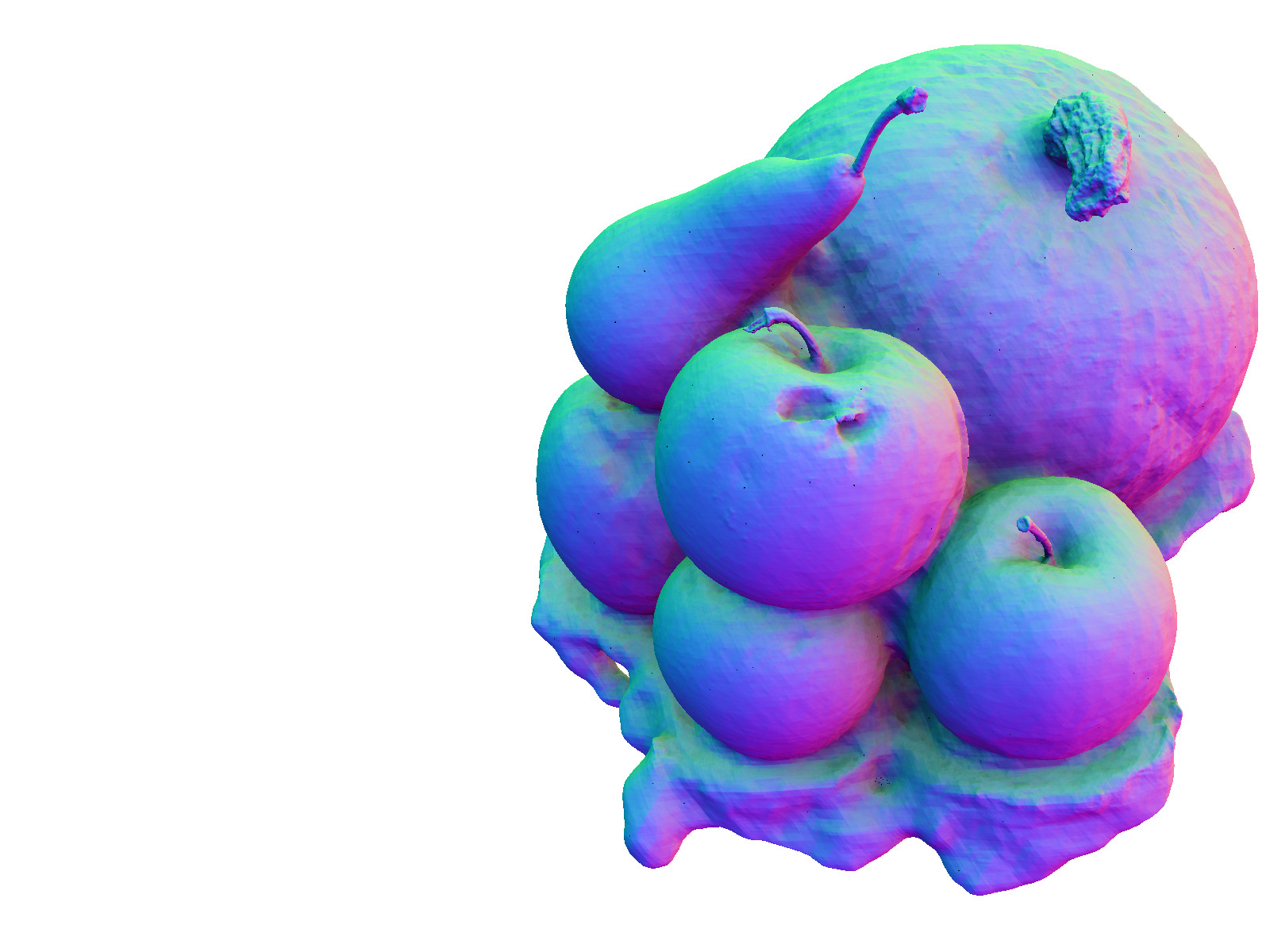}};
\node[inner sep=0pt] (Abl2) at (\ShiftX*2,0)
	{\includegraphics[
		trim=.4\WABL{} .13\HABL{} .0\WABL{} .0\HABL{},clip,
		width=.33\columnwidth, compress=false
		]{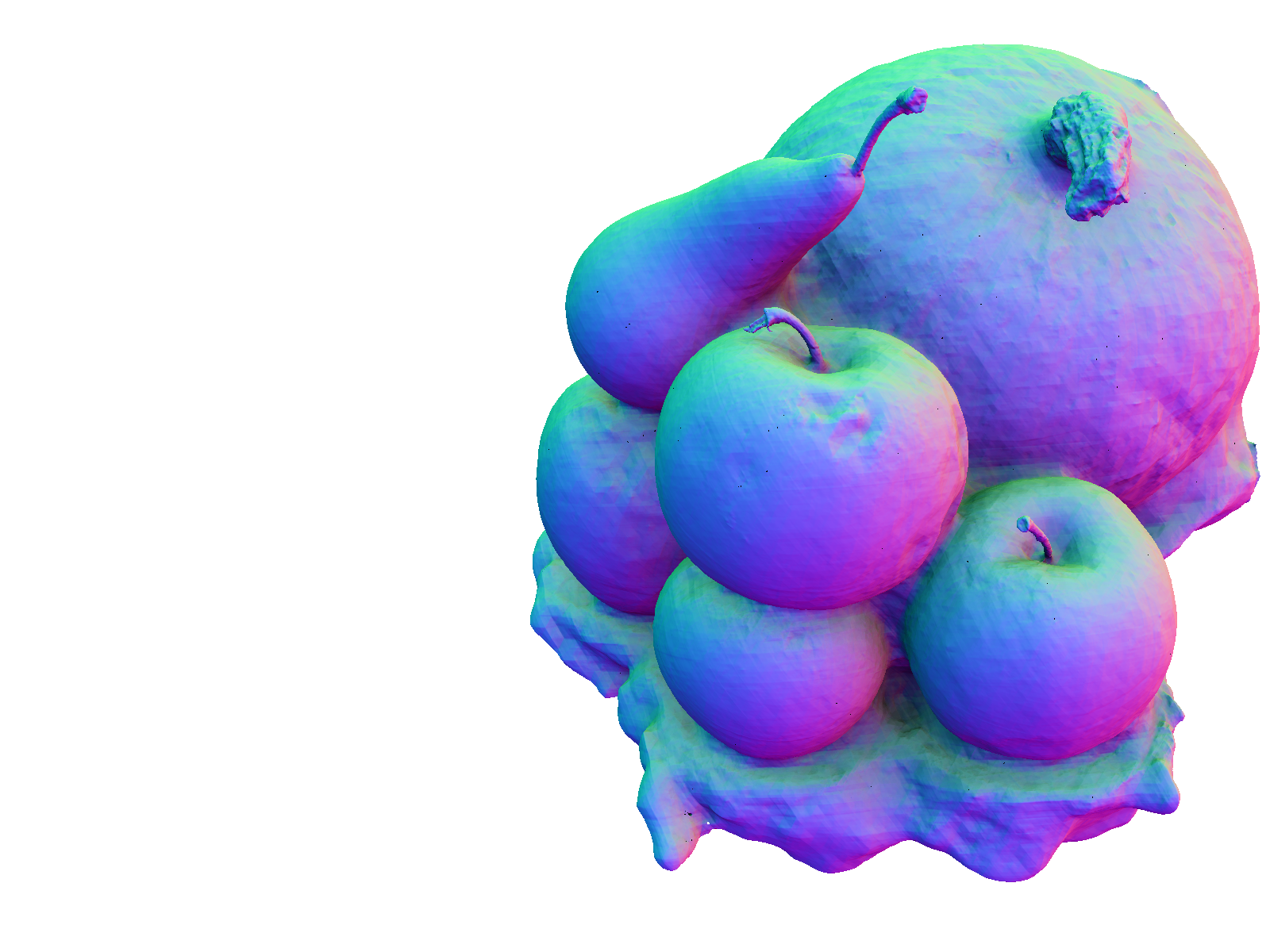}};
\node[inner sep=0pt] (Abl3) at (\ShiftX*3,0)
	{\includegraphics[
		trim=.4\WABL{} .13\HABL{} .0\WABL{} .0\HABL{},clip,
		width=.33\columnwidth, compress=false
		]{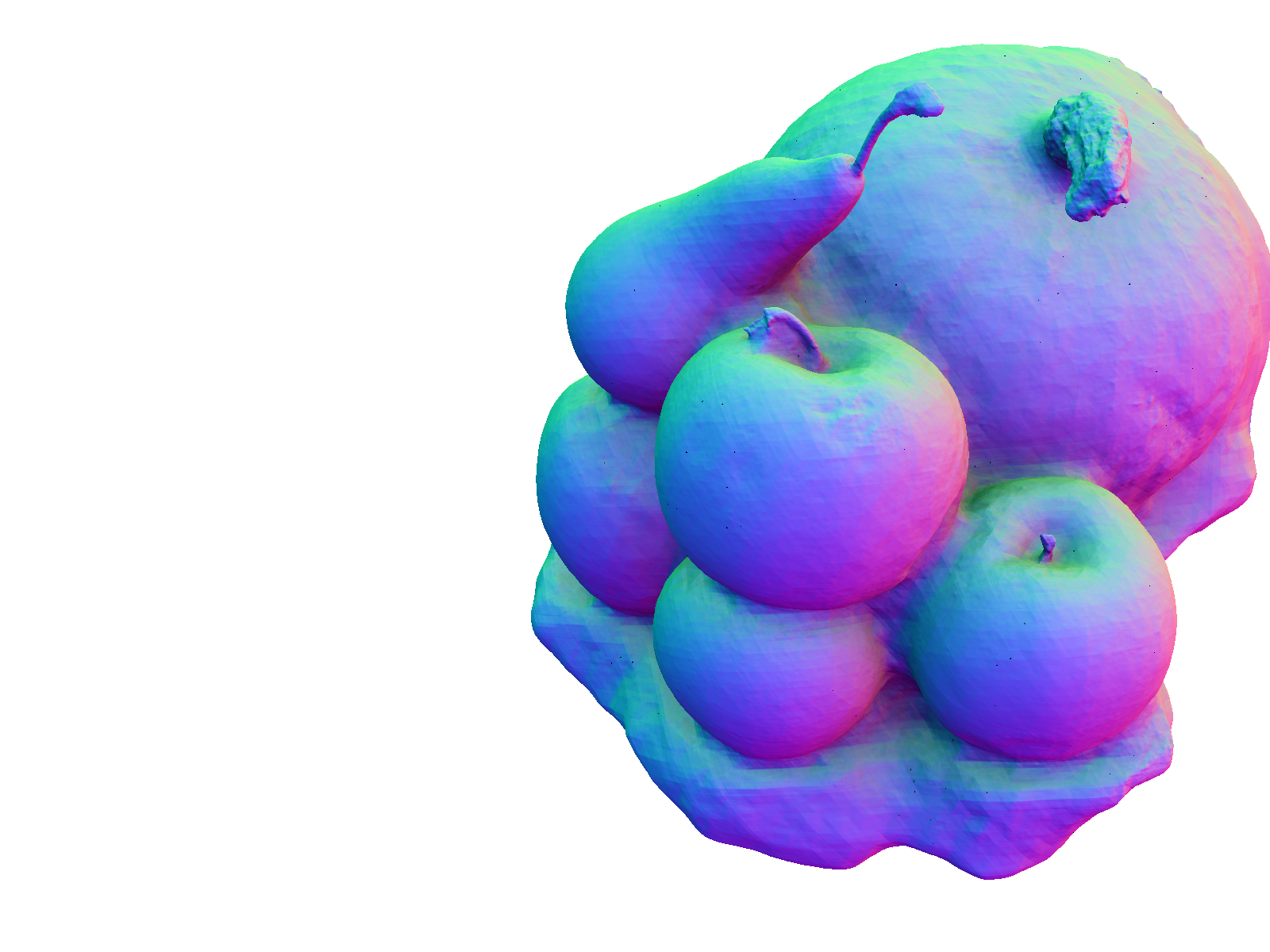}};
\node[inner sep=0pt] (Abl4) at (\ShiftX*4,0)
{\includegraphics[
	trim=.4\WABL{} .13\HABL{} .0\WABL{} .0\HABL{},clip,
	width=.33\columnwidth, compress=false
	]{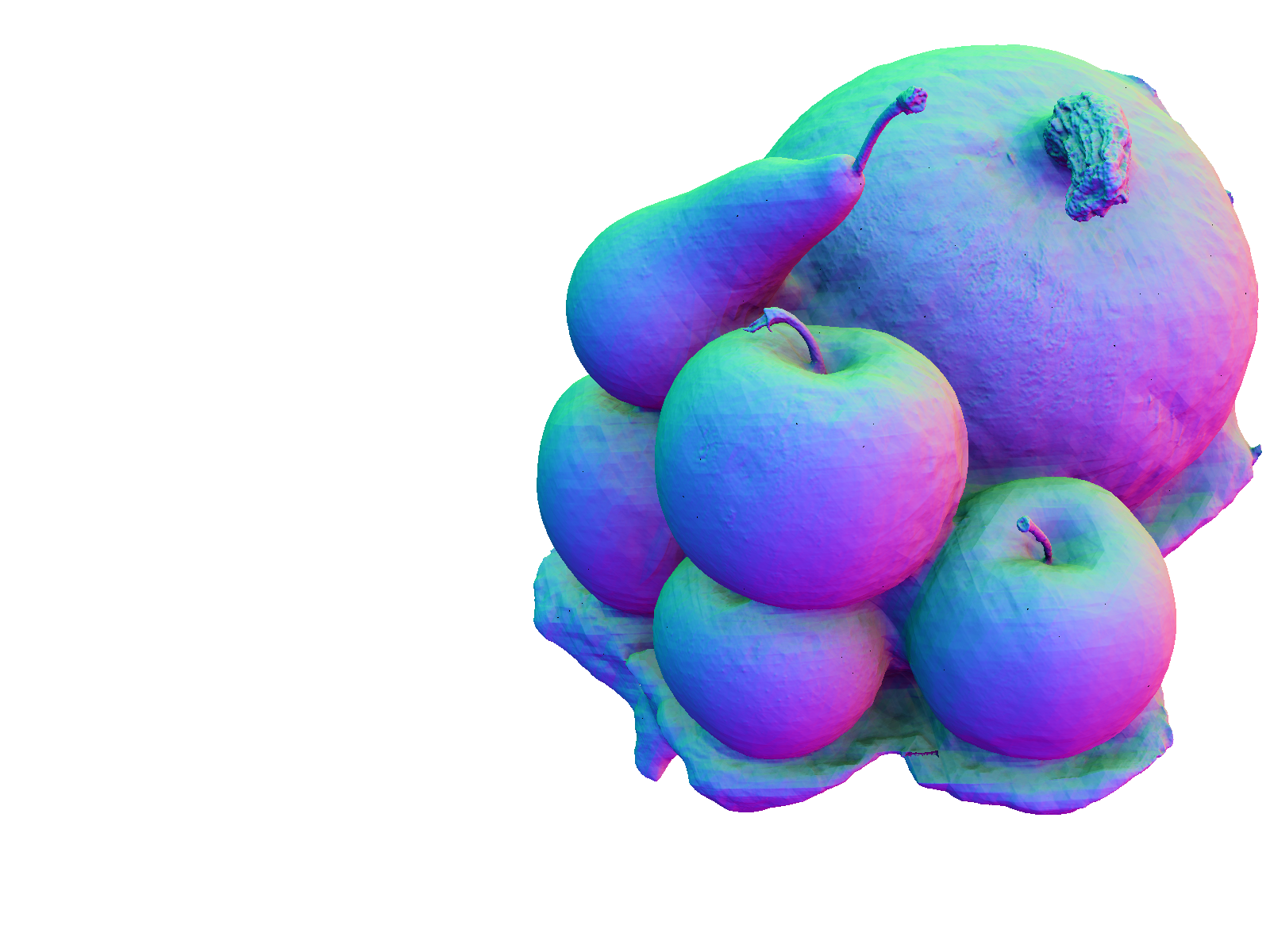}};
\node[inner sep=0pt] (GT) at (\ShiftX*5,0)
	{\includegraphics[
		trim=.4\WABL{} .13\HABL{} .0\WABL{} .0\HABL{},clip,
		width=.33\columnwidth, compress=false
		]{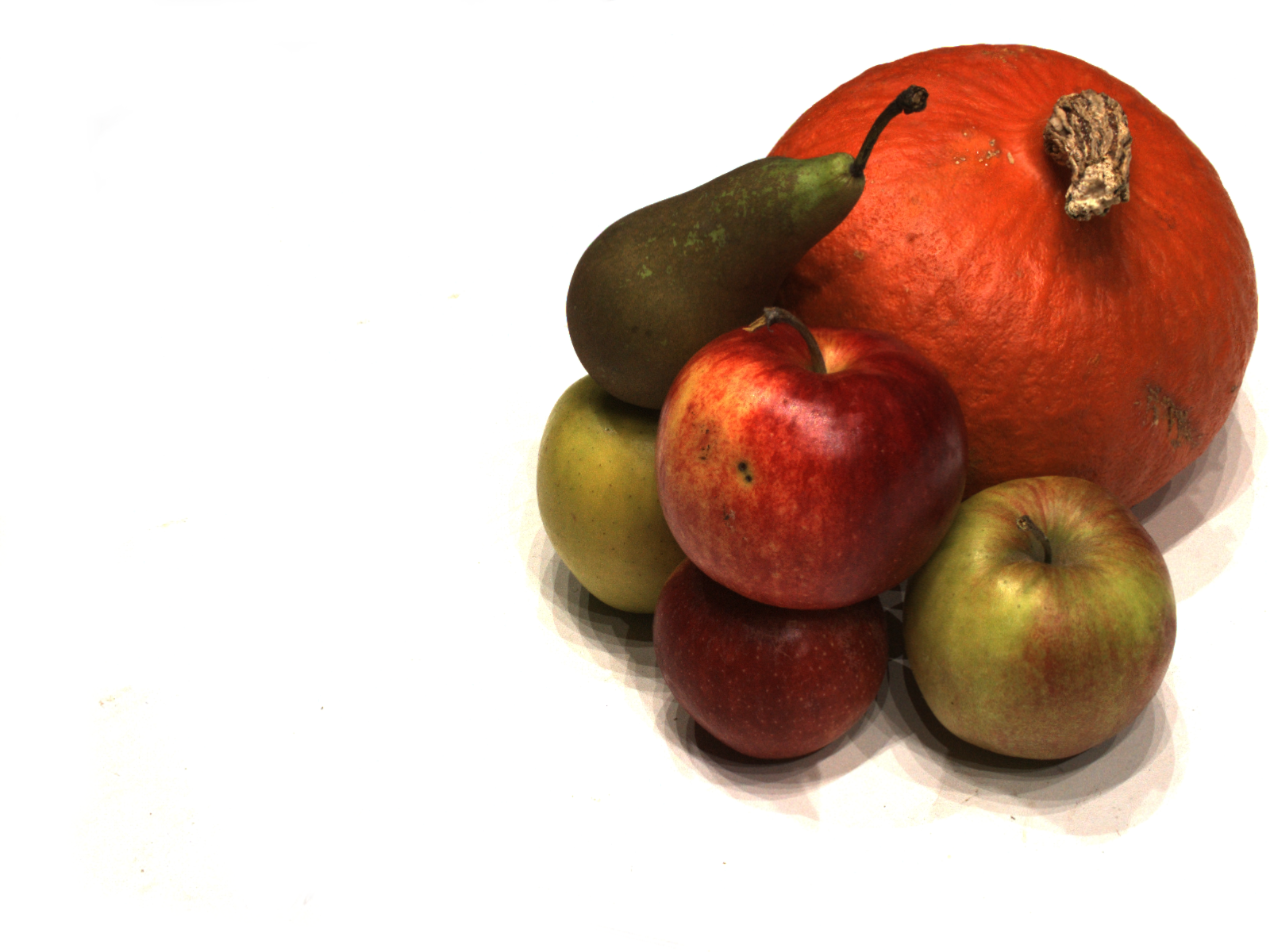}};

\node[] at (\ShiftX*0,-1.8) {\scriptsize Naive optimization };
\node[] at (\ShiftX*1,-1.8) {\scriptsize +Sphere init };
\node[] at (\ShiftX*2,-1.8) {\scriptsize +Coarse-to-fine };
\node[] at (\ShiftX*3,-1.8) {\scriptsize +Curvature loss };
\node[] at (\ShiftX*4,-1.8) {\scriptsize +RGB regularization };
\node[] at (\ShiftX*5,-1.8) {\scriptsize Reference image };

\draw [] (-1.5,-2.) -- (15.2,-2.);
\node[] at (\ShiftX*0,-2.2) {\scriptsize 0.950 };
\node[] at (\ShiftX*1,-2.2) {\scriptsize 1.032 };
\node[] at (\ShiftX*2,-2.2) {\scriptsize 1.012 };
\node[] at (\ShiftX*3,-2.2) {\scriptsize 1.189 };
\node[] at (\ShiftX*4,-2.2) {\scriptsize \bf{0.929} };
\node[] at (\ShiftX*5,-2.2) {\scriptsize Chamfer distance };

\spy [ph-orange,draw,height=1.2cm,width=1.2cm,magnification=2.5] on ($(Abl0.center) + (-0.2, 0.02)$) in node [line width=0.6mm, anchor=south west] at ($(Abl0.north west) + (0.15, 0.0) $);
\spy [ph-green,draw,height=1.2cm,width=1.2cm,magnification=4] on ($(Abl0.center) + (0.6, 1.0)$) in node [line width=0.6mm, anchor=south east] at ($(Abl0.north east) + (-0.15, 0.0) $);
\spy [ph-orange,draw,height=1.2cm,width=1.2cm,magnification=2.5] on ($(Abl1.center) + (-0.2, 0.02)$) in node [line width=0.6mm, anchor=south west] at ($(Abl1.north west) + (0.15, 0.0) $);
\spy [ph-green,draw,height=1.2cm,width=1.2cm,magnification=4] on ($(Abl1.center) + (0.6, 1.0)$) in node [line width=0.6mm, anchor=south east] at ($(Abl1.north east) + (-0.15, 0.0) $);
\spy [ph-orange,draw,height=1.2cm,width=1.2cm,magnification=2.5] on ($(Abl2.center) + (-0.2, 0.02)$) in node [line width=0.6mm, anchor=south west] at ($(Abl2.north west) + (0.15, 0.0) $);
\spy [ph-green,draw,height=1.2cm,width=1.2cm,magnification=4] on ($(Abl2.center) + (0.6, 1.0)$) in node [line width=0.6mm, anchor=south east] at ($(Abl2.north east) + (-0.15, 0.0) $);
\spy [ph-orange,draw,height=1.2cm,width=1.2cm,magnification=2.5] on ($(Abl3.center) + (-0.2, 0.02)$) in node [line width=0.6mm, anchor=south west] at ($(Abl3.north west) + (0.15, 0.0) $);
\spy [ph-green,draw,height=1.2cm,width=1.2cm,magnification=4] on ($(Abl3.center) + (0.6, 1.0)$) in node [line width=0.6mm, anchor=south east] at ($(Abl3.north east) + (-0.15, 0.0) $);
\spy [ph-orange,draw,height=1.2cm,width=1.2cm,magnification=2.5] on ($(Abl4.center) + (-0.2, 0.02)$) in node [line width=0.6mm, anchor=south west] at ($(Abl4.north west) + (0.15, 0.0) $);
\spy [ph-green,draw,height=1.2cm,width=1.2cm,magnification=4] on ($(Abl4.center) + (0.6, 1.0)$) in node [line width=0.6mm, anchor=south east] at ($(Abl4.north east) + (-0.15, 0.0) $);
\spy [ph-orange,draw,height=1.2cm,width=1.2cm,magnification=2.5] on ($(GT.center) + (-0.2, 0.02)$) in node [line width=0.6mm, anchor=south west] at ($(GT.north west) + (0.15, 0.0) $);
\spy [ph-green,draw,height=1.2cm,width=1.2cm,magnification=4] on ($(GT.center) + (0.6, 1.0)$) in node [line width=0.6mm, anchor=south east] at ($(GT.north east) + (-0.15, 0.0) $);

\end{tikzpicture}
\vspace{-2.5mm}
\caption{ Ablation study of the various components of our method. A naive combination of hash-based encoding and implicit surfaces can lead to undesirable holes and  overly smooth geometry.  Adding sphere initialization and coarse-to-fine optimization helps with recovering smoother surfaces especially for highly specular areas.  Adding a curvature loss helps further in remedying the issue of holes but uniformally smoothes the geometry. Adding RGB regularization forces the network to reconstruct the fine details.   } \label{fig:ablation_study}
\end{figure*} 
We perform an ablation study of the components that we propose for PermutoSDF in~\reffig{fig:ablation_study}. While sphere initialization and coarse-to-fine optimization help in recovering a smoother shape, they don't fully fix the issue of holes in the geometry. Adding the curvature loss solves most of the issues but also severely over-smoothes the object and results in a higher Chamfer distance. Adding RGB regularization via the Lipschitz loss is crucial to recover high-frequency details and obtaining the lowest Chamfer distance.

\subsection{4D Spatio-temporal Surface}
\begin{figure}
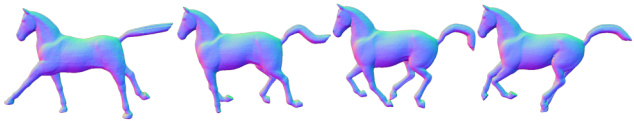

\centering
\small\sffamily
\begin{tikzpicture}

\newcommand\ShiftXLarge{2.0} %
\newcommand\ShiftYLarge{-0.4}

\node[inner sep=0pt] (d0) at (\ShiftXLarge*0,0)
    {\includegraphics[
    	height=.2\columnwidth
    	]{./imgs/4d_example/0_c.png}};
\node[inner sep=0pt] (d0) at (\ShiftXLarge*1+.2,0)
{\includegraphics[
	height=.2\columnwidth
	]{./imgs/4d_example/20_c.png}};
\node[inner sep=0pt] (d0) at (\ShiftXLarge*2+.2,0)
{\includegraphics[
	height=.2\columnwidth
	]{./imgs/4d_example/40_c.png}};
\node[inner sep=0pt] (d0) at (\ShiftXLarge*3+.2,0)
{\includegraphics[
	height=.2\columnwidth
	]{./imgs/4d_example/60_c.png}};

\end{tikzpicture}
\vspace{-7mm}
\caption{4D surface. Using our lattice we can efficiently learn surfaces that evolve in time. Here we visualize the learned geometry of a 4D model while we sweep through the time dimension.}  \label{fig:4d_example}
\vspace{-2mm}
\end{figure}

Since our lattice representation scales better in higher dimensions, we also include an experiment of encoding the surface of an object evolving through time as shown in~\reffig{fig:4d_example}.
 We directly fit the geometry of the 4D model by supervising with batches of orientated point samples from the surfaces of animated meshes. 
The loss function is similar to the one introduced in SIREN~\cite{siren} and we refer to the supplemental for more details. 
Our method successfully learned the evolving shape and can generate intermediate shapes by sweeping through time.

We note that learning 4D directly from images, similar to D-Nerf~\cite{dnerf}, is also possible. However, since the extension of the occupancy grid to 4D is not trivial and additional losses may also be needed to ensure smoothness in the time dimensions, we leave this for future work.

\section{Conclusion}
We proposed a combination of implicit surface representations and hash-based encoding methods for the task of reconstructing accurate geometry and appearance from unmasked posed color images. 

We improved upon the voxel-based hash encoding by using a permutohedral lattice which is always faster in training and faster for inference in three and higher dimensions. Additionally, we proposed a simple regularization scheme that allows to recover fine geometrical detail at the level of pores and wrinkles. Our full system can train in $\approx\SI{30}{\minute}$ on an RTX 3090 GPU and render in real-time using sphere tracing.
We believe this work together with the code release will help the community in a multitude of other tasks that require modeling signals using fast local features.

\vspace*{0.5ex}
\noindent \textbf{Acknowledgement.}
This work has been funded by the Deutsche Forschungsgemeinschaft (DFG, German Research Foundation) under Germany's Excellence Strategy - EXC 2070 - 390732324.

\balance

{\small
\bibliographystyle{ieee_fullname}
\bibliography{egbib}
}

\twocolumn[
\centering
{\bf\LARGE  Supplementary Material}
\vspace{20pt}
]

\section*{S1. Training Details}
For the first \SI{100}{\kilo{}} iterations, we train using the following loss function:
\begin{equation}
\Loss= \Loss_\textrm{rgb} + \lambda_{1} \Loss_\textrm{eik} + \lambda_2 \Loss_\textrm{curv},
\end{equation} 
where $\lambda_{1}=0.05$, $\lambda_{2}=1.5$.
For the remaining \SI{100}{\kilo{}} iteration, we remove $\lambda_2 \Loss_\textrm{curv}$ and replace it with $\lambda_3 \Loss_\textrm{Lipschitz}$,  where $\lambda_3=1\mathrm{e}{-5}$.

For 3D point sampling, we first create \num{64} uniform samples along each ray. We restrict the samples to be within the region that is defined as occupied by the occupancy grid. 
Afterwards, we run two iterations of importance sampling, each creating an additional \num{16} samples in the regions that are close to the surface. Concentrating samples close to the surface is crucial for recovering detail.

\section*{S2. Synthetic Data Comparison}
\begin{figure}[h]
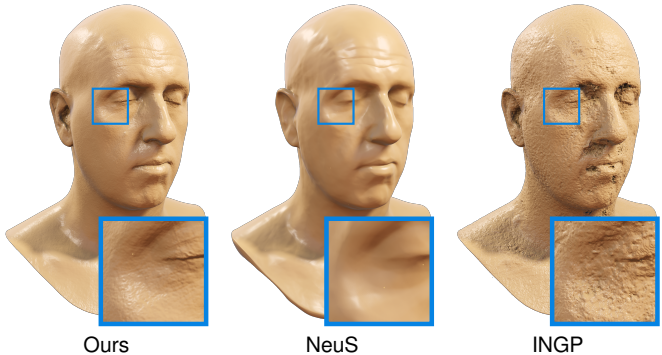

\centering
\small\sffamily
\begin{tikzpicture} [spy using outlines={ size=5cm,   every spy on node/.append style={thick}     }]

\newcommand\ShiftXLarge{3.0} %
\newcommand\ShiftySmall{1.1} %
\newcommand\ShiftyLarge{-4.5} %

\node[inner sep=0pt] (hash) at (\ShiftXLarge*0, 0)
    {\includegraphics[ width=.34\columnwidth, compress=false ]{./imgs/head_synth_comparison/hashsdf_v8_crop.png}};
\node[inner sep=0pt] (neus) at (\ShiftXLarge*1, 0)
{\includegraphics[ width=.34\columnwidth, compress=false 	]{./imgs/head_synth_comparison/neus_v8_crop.png}};
\node[inner sep=0pt] (ingp) at (\ShiftXLarge*2, 0)
{\includegraphics[ width=.34\columnwidth, compress=false ]{./imgs/head_synth_comparison/ingp_v8_crop.png}};

\node[] at (\ShiftXLarge*0,-2.4) {\footnotesize Ours };
\node[] at (\ShiftXLarge*1,-2.4) {\footnotesize NeuS };
\node[] at (\ShiftXLarge*2,-2.4) {\footnotesize INGP };

\spy [ph-blue,draw,height=1.4cm,width=1.4cm,magnification=3] on ($(hash.center) + (0.05, 0.77)$) in node [line width=0.6mm, anchor=south east] at ($(hash.south east) + (-0.1, 0.04) $);
\spy [ph-blue,draw,height=1.4cm,width=1.4cm,magnification=3] on ($(neus.center) + (0.05, 0.77)$) in node [line width=0.6mm, anchor=south east] at ($(neus.south east) + (-0.1, 0.04) $);
\spy [ph-blue,draw,height=1.4cm,width=1.4cm,magnification=3] on ($(ingp.center) + (0.05, 0.77)$) in node [line width=0.6mm, anchor=south east] at ($(ingp.south east) + (-0.1, 0.04) $);

\end{tikzpicture}
\caption{ We train our method, NeuS~\cite{neus}, and INGP~\cite{ingp} on the synthetically rendered dataset, described in Sec 7.3. We recover significantly more small detail than the other two methods. Best viewed zoomed-in. } \label{fig:head_synth_compare}
\end{figure}

We train also NeuS~\cite{neus} and INGP~\cite{ingp} on the synthetically rendered head dataset described in Sec 7.3. The recovered meshes are shown in~\reffig{fig:head_synth_compare}.

\section*{S3. Rendering Strategy}
\begin{figure}[bh!]
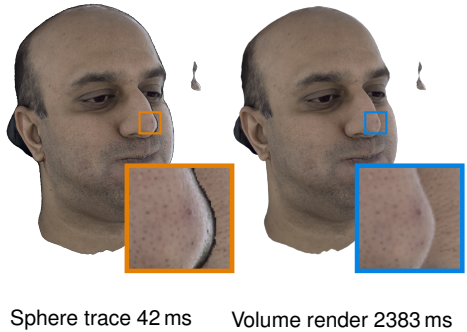

\centering
\small\sffamily
\begin{tikzpicture} [spy using outlines={ size=5cm,   every spy on node/.append style={thick}     }]

\newcommand\ShiftXLarge{3.0} %

\node[inner sep=0pt] (App) at (\ShiftXLarge*0, 0)
    {\includegraphics[ width=.44\columnwidth, compress=false ]{./imgs/vol_render_vs_sphere_trace/sphere_trace_10.png}};
\node[inner sep=0pt] (Geom) at (\ShiftXLarge*1, 0)
{\includegraphics[ width=.44\columnwidth, compress=false ]{./imgs/vol_render_vs_sphere_trace/vol_render.png}};

\spy [ph-orange,draw,height=1.4cm,width=1.4cm,magnification=5] on ($(App.center) + (0.65, 0.17)$) in node [line width=0.6mm, anchor=south east] at ($(App.south east) + (-0.1, 0.04) $);    
\spy [ph-blue,draw,height=1.4cm,width=1.4cm,magnification=5] on ($(Geom.center) + (0.65, 0.17)$) in node [line width=0.6mm, anchor=south east] at ($(Geom.south east) + (-0.04, 0.04) $);

\node at (0.0, -2.45) {\footnotesize  Sphere trace \SI{42}{\milli\second}  };
\node at (\ShiftXLarge+0.2,-2.45) {\footnotesize  Volume render  \SI{2383}{\milli\second} };

\end{tikzpicture}
\caption{ Sphere tracing is significantly faster than volumetric rendering, but it suffers from artifacts at surfaces with a grazing angle. } \label{fig:rendering_method}
\end{figure}

We compare images rendered through volumetric integration to the ones using sphere tracing. We observe that sphere tracing has the advantage of being significantly faster, as most rays converge towards the surface in few iterations. However, grazing surfaces require an arbitrary number of iterations and since we use a maximum of 20 iterations, these grazing surfaces may exhibit artifacts. 
A comparison between volumetric rendering and sphere tracing is shown in~\reffig{fig:rendering_method}.

\section*{S4. Tetrahedron vs Cube}
\begin{figure}
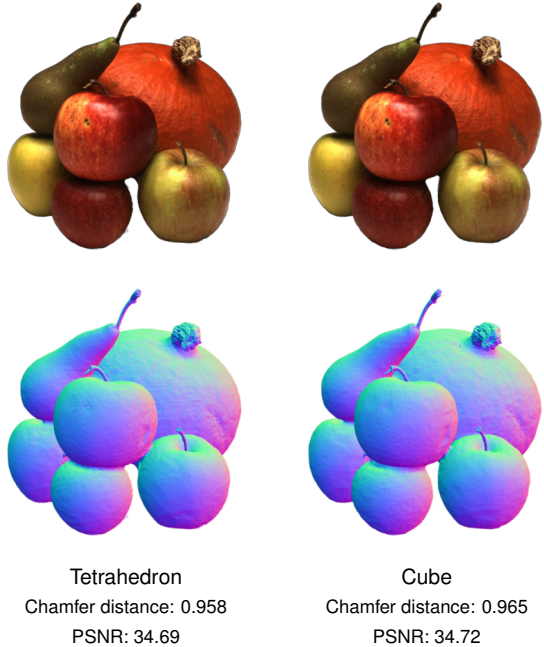

\centering
\small\sffamily
\begin{tikzpicture}

\newcommand\ShiftXLarge{4.0} %
\newcommand\ShiftySmall{3.8} %
\newcommand\ShiftyLarge{-4.5} %

\node[] at (0,0) {};%

\node[inner sep=0pt] (hash) at (\ShiftXLarge*0, 0)
{\includegraphics[ width=.45\columnwidth ]{./imgs/tetrahedron_vs_cube/tetrahedron/34rgb_c.png}};
\node[inner sep=0pt] (hash) at (\ShiftXLarge*1, 0)
{\includegraphics[ width=.45\columnwidth ]{./imgs/tetrahedron_vs_cube/cube/34rgb_c.png}};

\node[inner sep=0pt] (hash) at (\ShiftXLarge*0, -\ShiftySmall)
{\includegraphics[ width=.45\columnwidth ]{./imgs/tetrahedron_vs_cube/tetrahedron/34normals_c.png}};
\node[inner sep=0pt] (hash) at (\ShiftXLarge*1, -\ShiftySmall)
{\includegraphics[ width=.45\columnwidth ]{./imgs/tetrahedron_vs_cube/cube/34normals_c.png}};

\node[] at (\ShiftXLarge*0,-6.0) {\footnotesize Tetrahedron };
\node[] at (\ShiftXLarge*1,-6.0) {\footnotesize Cube };
\node[] at (\ShiftXLarge*0,-6.4) {\scriptsize Chamfer distance: 0.958 };
\node[] at (\ShiftXLarge*1,-6.4) {\scriptsize Chamfer distance: 0.965 };
\node[] at (\ShiftXLarge*0,-6.8) {\scriptsize PSNR: 34.69 };
\node[] at (\ShiftXLarge*1,-6.8) {\scriptsize PSNR: 34.72 };

\end{tikzpicture}
\caption{ We reconstruct the same scene using cube encoding and permutohedral encoding. We did not observe significant differences in the reconstruction quality. } \label{fig:tet_vs_cube}
\end{figure}

Apart from the speed improvements of using a permutohedral lattice instead of a hyper-cubical one, we are also interested on maintaining the encoding quality and therefore the reconstruction details. We reconstruct the same scene with both permutohedral encoding and cubical encoding as described in INGP~\cite{ingp}. 
We set the hash maps of both approaches to the same number of parameters, features per layer, and levels. We also extended the cubical lattice with the coarse-to-fine optimization in order match the optimization behavior of the permutohedral lattice.
In~\reffig{fig:tet_vs_cube}, we show both reconstructions and compare their Chamfer distance and PSNR values for novel-view synthesis. We did not observe a significant difference in the reconstruction quality. 
\section*{S5. Occupancy Grid Update}
We initialize an occupancy grid with all the voxels being occupied and with an initial SDF that is constant zero. This ensures that we sample everywhere at the beginning of training.

For updating the occupancy grid, we use the following steps:
\begin{itemize}
	\setlength\itemsep{0.01em}
	\item Every 8th iteration of training, we sample $2^{18}$ random points within the bounding box that contains the scene.
	\item We obtain the SDF value $s_\pos$ for each point $\pos$ by running a forward pass through the model.
	\item We obtain the old SDF value $s_{old}$ stored for the voxel in which the point falls into.
	\item We compute a new SDF value for this voxel $s_{new}$ as the exponential average of the old SDF for the voxel and the SDF for the point: $s_{new} = s_{old} + 0.3(s_\pos-s_{old})$.
	\item Since we are discretizing the SDF to a grid, and we don't want to miss any possible low SDF values that we would want to sample, we compute the minimum possible SDF that can be reached within this voxel under the assumption of perfect Eikonal loss. For this, we use: $s_{min} = max(0, \abs{s_{new}} - d) $, where $d$ is the length of the voxel diagonal.
	\item Using the logistic density distribution as described in NeuS~\cite{neus}, we compute the weight that this sample would contribute to the volumetric render---assuming no obstruction from other samples:\\ $w = \SigmoidSlope \cdot e^{-\SigmoidSlope \cdot s_{min}}/(1 + e^{-\SigmoidSlope \cdot s_{min}})^2$. 
	\item If the weight $w$ falls bellow a specified threshold, we set the voxel to unoccupied and therefore don't create samples within it anymore.
\end{itemize}

\section*{S6. Color Calibration}
We observe that some datasets exhibit images with different exposure times. This discrepancy between images can influence both the reconstruction and the obtained color field as the network would try to explain the variability with view-dependent effects.  We circumvent this by learning a per-camera gain $g=(1+\Delta g)$ and bias $b$ so that the reconstructed color for each camera is $c=\sigma( \hat{c} \cdot g + b)$, where $\hat{c}$ is the raw color output from the network and $\sigma$ is a sigmoid function that restricts the color to the correct range. 
We set a selected camera (usually the first one from the dataset) to have $g=1$ and $b=0$ and apply weight decay to $\Delta g$ and $b$ to further ensure that the calibration doesn't alter the colors unnecessarily.

\section*{S7. 4D Spatial-temporal Surface}
For fitting a 4D surface, we sample random points from animated 3D meshes. These 3D points are concatenated with a time dimension that ranges from 0 to 1, where 0 is the start time of the animation and 1 is the end. We define these 4D samples at the surface of the mesh as $\pos_s$. We also compute the normal $\mathbf{n}_s$ for each on the surface samples.
We additionally define random 4D samples in a bounded domain around the animated mesh which we denote with $\pos_r$ 
Afterwards, we learn a model $g(\encOut; \NetParams)$ together with an encoding ${\encOut = \enc(\pos; \LatticeParams)}$ that maps from the 4D coordinate to an SDF value. 
For this, we follow the approach of SIREN~\cite{siren} and use a loss of the form:

\begin{equation}
\begin{split}
\Loss_\textrm{sdf} &= \sum_{\pos_s \cup \pos_r}  \left(  \norm{  \nabla g(\enc(\pos))    }  -1   \right)^2 \\
&+ \sum_{\pos_s}  \norm{g(\enc(\pos)} \\
&+ \sum_{\pos_s}  (\nabla g(\enc(\pos)) \cdot \mathbf{n}_s  -1) \\
&+ \sum_{\pos_r} exp(-\alpha \cdot \abs{g(\enc(\pos))}).
\end{split}
\end{equation}

In this 4D experiment, no explicit smoothness was enforced in the temporal domain since we didn't find it necessary. We sample from an animation of 100 frames so the temporal resolution is relatively high. At lower temporal resolution, smoothness might again become a concern.

Nevertheless, this approach shows that our model can deal with 4D representations onto which further ideas, like dynamic deformation fields, can be built upon.

\section*{S8. Number of Cameras}

In order to study the robustness of our method to the number of input images, we vary the number of images used for reconstruction as shown in~\reffig{fig:nr_cameras}. Due to the curvature loss, our method can recover smooth but plausible surfaces even with as low as seven input images. However, for a smaller number of input images we observe a high likelihood of not converging to the correct surface.

\begin{figure}
\centering
\small\sffamily
\begin{tikzpicture} [spy using outlines={ size=5cm,   every spy on node/.append style={thick}     }]

\newcommand\ShiftXLarge{1.7} %
\newcommand\ShiftySmall{1.1} %
\newcommand\ShiftyLarge{-4.5} %

\node[inner sep=0pt] (hash) at (\ShiftXLarge*0, 0)
    {\includegraphics[ width=.2\columnwidth ]{./imgs/ablation_nr_images/3_c.png}};
\node[inner sep=0pt] (hash) at (\ShiftXLarge*1, 0)
	{\includegraphics[ width=.2\columnwidth ]{./imgs/ablation_nr_images/7_c.png}};
\node[inner sep=0pt] (hash) at (\ShiftXLarge*2, 0)
	{\includegraphics[ width=.2\columnwidth ]{./imgs/ablation_nr_images/14_c.png}};
\node[inner sep=0pt] (hash) at (\ShiftXLarge*3, 0)
	{\includegraphics[ width=.2\columnwidth ]{./imgs/ablation_nr_images/28_c.png}};
\node[inner sep=0pt] (hash) at (\ShiftXLarge*4, 0)
{\includegraphics[ width=.2\columnwidth ]{./imgs/ablation_nr_images/56_c.png}};

\node[align=left] at (\ShiftXLarge*0,-1.6) {\scriptsize 3 };
\node[align=left] at (\ShiftXLarge*1,-1.6) {\scriptsize 7  };
\node[align=left] at (\ShiftXLarge*2,-1.6) {\scriptsize 14 };
\node[align=left] at (\ShiftXLarge*3,-1.6) {\scriptsize 28 };
\node[align=left] at (\ShiftXLarge*4,-1.6) {\scriptsize 56 };

\end{tikzpicture}
\caption{ We experiment with the number of input images for our method. We observe significant degradation at around 7 input images and a failure to converge at 3 images. } \label{fig:nr_cameras}
\end{figure} 
\section*{S9. Training schedule}
We follow a fixed training schedule over \num{200}k iterations. This includes a phase where we train with curvature loss in order to recover the rough shape, and another phase with RGB regularization to recover detail. In order to study the robustness of our method to this schedule, we expand and contract the schedule to be as long as \num{300}k iteration or as short as \num{50}k iterations and show the results in~\reffig{fig:ablation_schedule}. By modifying the schedule, we proportionally expand or contract the time that is spent optimizing the sphere, training with high curvature, and training with RGB regularization. We observe that the model is quite robust to different schedules and only for the very short ones it fails to recover some of the geometry. In general, we found that view-dependent effects like the highlight on the apple are the parts that take the longest to converge to good geometry. Most objects are reconstructed well with shorter schedules but our default schedule of \num{200}k iteration is a good trade-off between optimization speed and accuracy. 

\begin{figure}
\centering
\small\sffamily
\begin{tikzpicture} [spy using outlines={ size=5cm,   every spy on node/.append style={thick}     }]

\newcommand\ShiftXLarge{1.7} %
\newcommand\ShiftySmall{1.1} %
\newcommand\ShiftyLarge{-4.5} %

\node[inner sep=0pt] (hash) at (\ShiftXLarge*0, 0)
    {\includegraphics[ width=.19\columnwidth ]{./imgs/ablation_schedule/0p25_c.png}};
\node[inner sep=0pt] (hash) at (\ShiftXLarge*1, 0)
	{\includegraphics[ width=.19\columnwidth ]{./imgs/ablation_schedule/0p5_c.png}};
\node[inner sep=0pt] (hash) at (\ShiftXLarge*2, 0)
	{\includegraphics[ width=.19\columnwidth ]{./imgs/ablation_schedule/0p75_c.png}};
\node[inner sep=0pt] (hash) at (\ShiftXLarge*3, 0)
	{\includegraphics[ width=.19\columnwidth ]{./imgs/ablation_schedule/1_c.png}};
\node[inner sep=0pt] (hash) at (\ShiftXLarge*4, 0)
	{\includegraphics[ width=.19\columnwidth ]{./imgs/ablation_schedule/1p5_c.png}};

\node[align=left] at (\ShiftXLarge*0,-1.1) {\scriptsize \num{50}k };
\node[align=left] at (\ShiftXLarge*1,-1.1) {\scriptsize \num{100}k  };
\node[align=left] at (\ShiftXLarge*2,-1.1) {\scriptsize \num{150}k };
\node[align=left] at (\ShiftXLarge*3,-1.1) {\scriptsize \num{200}k (default) };
\node[align=left] at (\ShiftXLarge*4,-1.1) {\scriptsize \num{300}k };

\end{tikzpicture}
\caption{ We follow a fixed training schedule that finishes after \num{200}k iteration. We expand and contract this fixed schedule to be shorter or longer in order to test robustness. We see that for a schedule of \num{50}k the method fails to reconstruct the geometry for the highlight of the apple. Our default of \num{200}k can recover good geometry in reasonable time. A longer schedule results in better reconstructions at the cost of more optimization time. } \label{fig:ablation_schedule}
\end{figure} 
\vspace{-1pt}
\section*{S10. Reflective Surfaces}

\begin{figure}
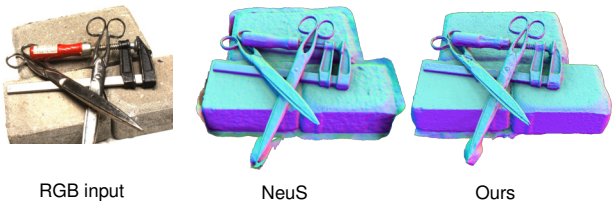

\centering
\small\sffamily
\begin{tikzpicture} [spy using outlines={ size=5cm,   every spy on node/.append style={thick}     }]

\newcommand\ShiftXLarge{2.7} %
\newcommand\ShiftySmall{1.1} %
\newcommand\ShiftyLarge{-4.5} %

\node[inner sep=0pt] (hash) at (\ShiftXLarge*0, 0)
{\raisebox{2ex}{\includegraphics[ width=.28\columnwidth ]{./imgs/reflective_surface/000014_cut.png}}};
\node[inner sep=0pt] (hash) at (\ShiftXLarge*1, 0)
    {\includegraphics[ width=.384\columnwidth ]{./imgs/reflective_surface/neus_64.png}};
\node[inner sep=0pt] (hash) at (\ShiftXLarge*2+0.1, 0)
	{\includegraphics[ width=.384\columnwidth ]{./imgs/reflective_surface/permuto_sdf_64.png}};

   \node[align=left] at (\ShiftXLarge*0,-1.6) {\scriptsize RGB input };    
\node[align=left] at (\ShiftXLarge*1,-1.6) {\scriptsize NeuS };
\node[align=left] at (\ShiftXLarge*2+0.1,-1.6) {\scriptsize Ours  };

\end{tikzpicture}
\caption{ We observe that our model sometimes struggle with very reflective surface like the metal on the scissors. It tends to add noisy geometry to these surfaces in order to explain the view-dependent effects. Object priors or a higher curvature loss for this kind of objects could alleviate the issue.} \label{fig:reflective_surfaces}
\end{figure}

Our model tries to explain large color changes with changes in geometry. This behavior can be detrimental in the case of mirror-like surfaces. As we observe in~\reffig{fig:reflective_surfaces}, NeuS recovers a smooth surface on the metal scissors while our method exhibits noise. This can be seen as a general limitation of RGB reconstruction methods where it is difficult to know if changes in color are from view-dependent effects or from high-frequency geometry. A model that learns object priors might perform better in these cases.

\vspace{-1pt}
\section*{S11. Thin Structures}

\begin{figure}
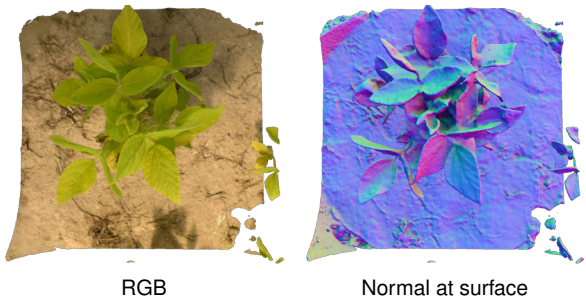

\centering
\small\sffamily
\begin{tikzpicture}

\newcommand\ShiftXLarge{4.0} %
\newcommand\ShiftySmall{3.8} %
\newcommand\ShiftyLarge{-4.5} %

\node[] at (0,0) {};%

\node[inner sep=0pt] (hash) at (\ShiftXLarge*0, 0)
{\includegraphics[ width=.45\columnwidth ]{./imgs/plant/rgb1_c.png}};
\node[inner sep=0pt] (hash) at (\ShiftXLarge*1, 0)
{\includegraphics[ width=.45\columnwidth ]{./imgs/plant/normal1_c.png}};

\node[] at (\ShiftXLarge*0,-2.0) {\footnotesize RGB };
\node[] at (\ShiftXLarge*1,-2.0) {\footnotesize Normal at surface };

\end{tikzpicture}
\caption{ Plant reconstruction is an especially difficult case since it features many self-occlusions and thin structures. We observe that our method can deal well with this kind of data despite using only 14 images as input. } \label{fig:plant}
\end{figure} 
An interesting case to test for our SDF-based method is reconstructing thin structures. For this, we capture 14 images of a plant with relatively complex geometry with many leafs and self occlusions. \reffig{fig:plant} shows a rendered novel view and surface normals. Our method reconstructs accurate color and plausible geometry. Despite some errors that are to be expected given the low image count, it can recover thin steams and leaves which shows that our method is robust to this kind of data. 

\vspace{-1pt}
\section*{S12. Qualitative DTU Results}

\begin{figure*}[]
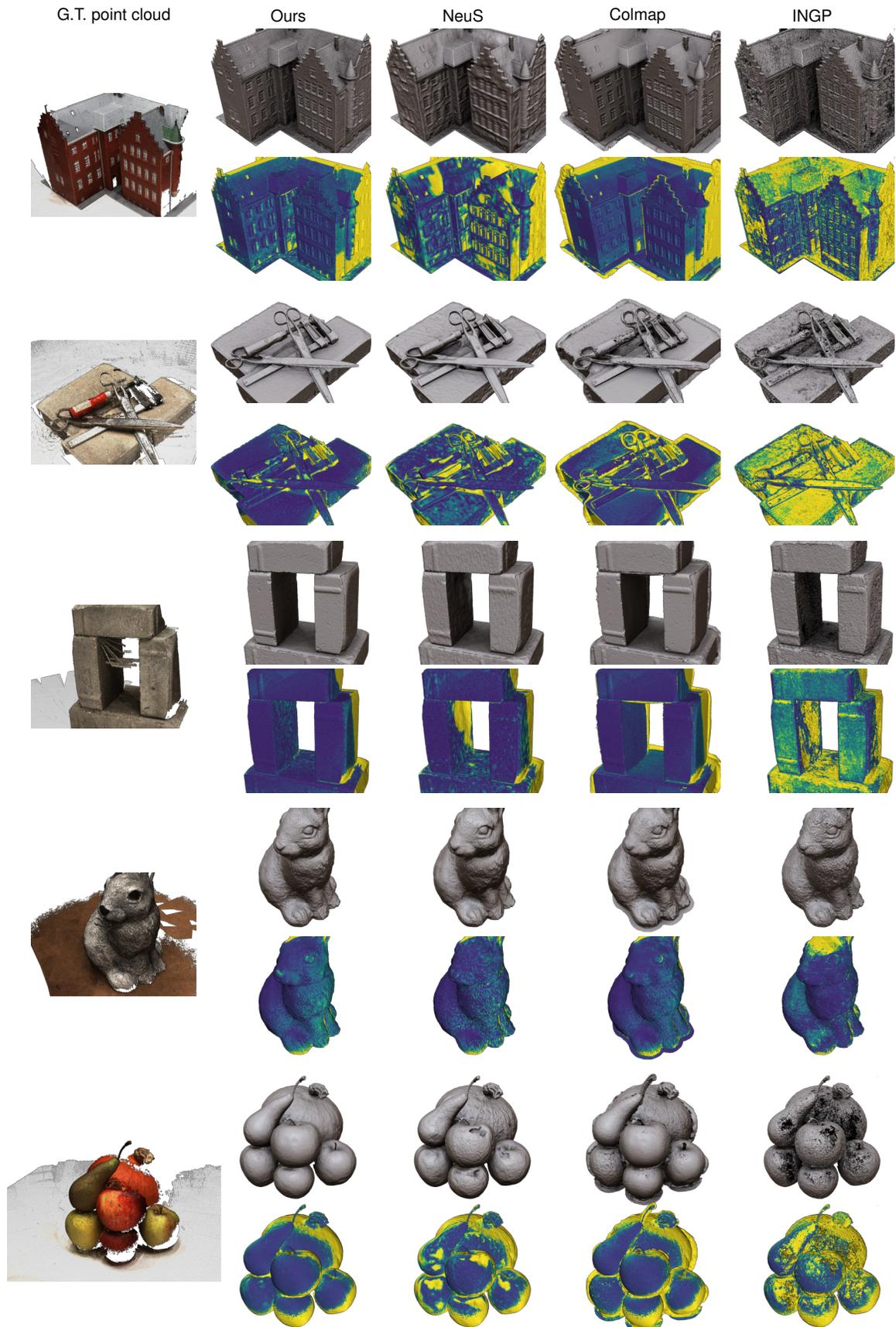

\centering
\small\sffamily
\begin{tikzpicture}

\newcommand\ShiftXLarge{3.0} %
\newcommand\ShiftySmall{1.1} %
\newcommand\ShiftyLarge{-4.5} %

\node[inner sep=0pt] (d0) at (\ShiftXLarge*0, \ShiftyLarge*0)
    {\includegraphics[ width=.34\columnwidth ]{./imgs/dtu_comparison_mesh/dtu_scan24/gt_cloud/23.png}};
\node[inner sep=0pt] (d0) at (\ShiftXLarge*1, \ShiftySmall +\ShiftyLarge*0)
{\includegraphics[ width=.34\columnwidth ]{./imgs/dtu_comparison_mesh/dtu_scan24/mine/23.png}};
\node[inner sep=0pt] (d0) at (\ShiftXLarge*2, \ShiftySmall +\ShiftyLarge*0)
{\includegraphics[ width=.34\columnwidth ]{./imgs/dtu_comparison_mesh/dtu_scan24/neus/23.png}};
\node[inner sep=0pt] (d0) at (\ShiftXLarge*3, \ShiftySmall +\ShiftyLarge*0)
{\includegraphics[ width=.34\columnwidth ]{./imgs/dtu_comparison_mesh/dtu_scan24/colmap/23.png}};
\node[inner sep=0pt] (d0) at (\ShiftXLarge*4, \ShiftySmall +\ShiftyLarge*0)
{\includegraphics[ width=.34\columnwidth ]{./imgs/dtu_comparison_mesh/dtu_scan24/ingp/23.png}};
\node[inner sep=0pt] (d0) at (\ShiftXLarge*1, -\ShiftySmall +\ShiftyLarge*0)
{\includegraphics[ width=.34\columnwidth ]{./imgs/dtu_comparison_error/dtu_scan24/mine/23.png}};
\node[inner sep=0pt] (d0) at (\ShiftXLarge*2, -\ShiftySmall +\ShiftyLarge*0)
{\includegraphics[ width=.34\columnwidth ]{./imgs/dtu_comparison_error/dtu_scan24/neus/23.png}};
\node[inner sep=0pt] (d0) at (\ShiftXLarge*3, -\ShiftySmall +\ShiftyLarge*0)
{\includegraphics[ width=.34\columnwidth ]{./imgs/dtu_comparison_error/dtu_scan24/colmap/23.png}};
\node[inner sep=0pt] (d0) at (\ShiftXLarge*4, -\ShiftySmall +\ShiftyLarge*0)
{\includegraphics[ width=.34\columnwidth ]{./imgs/dtu_comparison_error/dtu_scan24/ingp/23.png}};

\node[inner sep=0pt] (d0) at (\ShiftXLarge*0, \ShiftyLarge*1 +.25)
{\includegraphics[ width=.34\columnwidth ]{./imgs/dtu_comparison_mesh/dtu_scan37/gt_cloud/12.png}};
\node[inner sep=0pt] (d0) at (\ShiftXLarge*1, \ShiftySmall +\ShiftyLarge*1 +.2)
{\includegraphics[ width=.34\columnwidth ]{./imgs/dtu_comparison_mesh/dtu_scan37/mine/12.png}};
\node[inner sep=0pt] (d0) at (\ShiftXLarge*2, \ShiftySmall +\ShiftyLarge*1 +.2)
{\includegraphics[ width=.34\columnwidth ]{./imgs/dtu_comparison_mesh/dtu_scan37/neus/12.png}};
\node[inner sep=0pt] (d0) at (\ShiftXLarge*3, \ShiftySmall +\ShiftyLarge*1 +.2)
{\includegraphics[ width=.34\columnwidth ]{./imgs/dtu_comparison_mesh/dtu_scan37/colmap/12.png}};
\node[inner sep=0pt] (d0) at (\ShiftXLarge*4, \ShiftySmall +\ShiftyLarge*1 +.2)
{\includegraphics[ width=.34\columnwidth ]{./imgs/dtu_comparison_mesh/dtu_scan37/ingp/12.png}};

\node[inner sep=0pt] (d0) at (\ShiftXLarge*1, -\ShiftySmall +\ShiftyLarge*1 +.3)
{\includegraphics[ width=.34\columnwidth ]{./imgs/dtu_comparison_error/dtu_scan37/mine/12.png}};
\node[inner sep=0pt] (d0) at (\ShiftXLarge*2, -\ShiftySmall +\ShiftyLarge*1 +.3)
{\includegraphics[ width=.34\columnwidth ]{./imgs/dtu_comparison_error/dtu_scan37/neus/12.png}};
\node[inner sep=0pt] (d0) at (\ShiftXLarge*3, -\ShiftySmall +\ShiftyLarge*1 +.3)
{\includegraphics[ width=.34\columnwidth ]{./imgs/dtu_comparison_error/dtu_scan37/colmap/12.png}};
\node[inner sep=0pt] (d0) at (\ShiftXLarge*4, -\ShiftySmall +\ShiftyLarge*1 +.3)
{\includegraphics[ width=.34\columnwidth ]{./imgs/dtu_comparison_error/dtu_scan37/ingp/12.png}};

\node[inner sep=0pt] (d0) at (\ShiftXLarge*0, \ShiftyLarge*2 +.2)
{\includegraphics[ width=.34\columnwidth ]{./imgs/dtu_comparison_mesh/dtu_scan40/gt_cloud/35.png}};
\node[inner sep=0pt] (d0) at (\ShiftXLarge*1, \ShiftySmall +\ShiftyLarge*2 +.2)
{\includegraphics[ width=.34\columnwidth ]{./imgs/dtu_comparison_mesh/dtu_scan40/mine/35.png}};
\node[inner sep=0pt] (d0) at (\ShiftXLarge*2, \ShiftySmall +\ShiftyLarge*2 +.2)
{\includegraphics[ width=.34\columnwidth ]{./imgs/dtu_comparison_mesh/dtu_scan40/neus/35.png}};
\node[inner sep=0pt] (d0) at (\ShiftXLarge*3, \ShiftySmall +\ShiftyLarge*2 +.2)
{\includegraphics[ width=.34\columnwidth ]{./imgs/dtu_comparison_mesh/dtu_scan40/colmap/35.png}};
\node[inner sep=0pt] (d0) at (\ShiftXLarge*4, \ShiftySmall +\ShiftyLarge*2 +.2)
{\includegraphics[ width=.34\columnwidth ]{./imgs/dtu_comparison_mesh/dtu_scan40/ingp/35.png}};
\node[inner sep=0pt] (d0) at (\ShiftXLarge*1, -\ShiftySmall +\ShiftyLarge*2 +.2)
{\includegraphics[ width=.34\columnwidth ]{./imgs/dtu_comparison_error/dtu_scan40/mine/35.png}};
\node[inner sep=0pt] (d0) at (\ShiftXLarge*2, -\ShiftySmall +\ShiftyLarge*2 +.2)
{\includegraphics[ width=.34\columnwidth ]{./imgs/dtu_comparison_error/dtu_scan40/neus/35.png}};
\node[inner sep=0pt] (d0) at (\ShiftXLarge*3, -\ShiftySmall +\ShiftyLarge*2 +.2)
{\includegraphics[ width=.34\columnwidth ]{./imgs/dtu_comparison_error/dtu_scan40/colmap/35.png}};
\node[inner sep=0pt] (d0) at (\ShiftXLarge*4, -\ShiftySmall +\ShiftyLarge*2 +.2)
{\includegraphics[ width=.34\columnwidth ]{./imgs/dtu_comparison_error/dtu_scan40/ingp/35.png}};

\node[inner sep=0pt] (d0) at (\ShiftXLarge*0, \ShiftyLarge*3 +.1)
{\includegraphics[ width=.34\columnwidth ]{./imgs/dtu_comparison_mesh/dtu_scan55/gt_cloud/17.png}};
\node[inner sep=0pt] (d0) at (\ShiftXLarge*1, \ShiftySmall +\ShiftyLarge*3+.1)
{\includegraphics[ width=.34\columnwidth ]{./imgs/dtu_comparison_mesh/dtu_scan55/mine/17.png}};
\node[inner sep=0pt] (d0) at (\ShiftXLarge*2, \ShiftySmall +\ShiftyLarge*3 +.1)
{\includegraphics[ width=.34\columnwidth ]{./imgs/dtu_comparison_mesh/dtu_scan55/neus/17.png}};
\node[inner sep=0pt] (d0) at (\ShiftXLarge*3, \ShiftySmall +\ShiftyLarge*3 +.1)
{\includegraphics[ width=.34\columnwidth ]{./imgs/dtu_comparison_mesh/dtu_scan55/colmap/17.png}};
\node[inner sep=0pt] (d0) at (\ShiftXLarge*4, \ShiftySmall +\ShiftyLarge*3 +.1)
{\includegraphics[ width=.34\columnwidth ]{./imgs/dtu_comparison_mesh/dtu_scan55/ingp/17.png}};
\node[inner sep=0pt] (d0) at (\ShiftXLarge*1, -\ShiftySmall +\ShiftyLarge*3 +.1)
{\includegraphics[ width=.34\columnwidth ]{./imgs/dtu_comparison_error/dtu_scan55/mine/17.png}};
\node[inner sep=0pt] (d0) at (\ShiftXLarge*2, -\ShiftySmall +\ShiftyLarge*3 +.1)
{\includegraphics[ width=.34\columnwidth ]{./imgs/dtu_comparison_error/dtu_scan55/neus/17.png}};
\node[inner sep=0pt] (d0) at (\ShiftXLarge*3, -\ShiftySmall +\ShiftyLarge*3 +.1)
{\includegraphics[ width=.34\columnwidth ]{./imgs/dtu_comparison_error/dtu_scan55/colmap/17.png}};
\node[inner sep=0pt] (d0) at (\ShiftXLarge*4, -\ShiftySmall +\ShiftyLarge*3 +.1)
{\includegraphics[ width=.34\columnwidth ]{./imgs/dtu_comparison_error/dtu_scan55/ingp/17.png}};

\node[inner sep=0pt] (d0) at (\ShiftXLarge*0, \ShiftyLarge*4)
{\includegraphics[ width=.44\columnwidth ]{./imgs/dtu_comparison_mesh/dtu_scan63/gt_cloud/24.png}};
\node[inner sep=0pt] (d0) at (\ShiftXLarge*1, \ShiftySmall +\ShiftyLarge*4)
{\includegraphics[ width=.44\columnwidth ]{./imgs/dtu_comparison_mesh/dtu_scan63/mine/24.png}};
\node[inner sep=0pt] (d0) at (\ShiftXLarge*2, \ShiftySmall +\ShiftyLarge*4)
{\includegraphics[ width=.44\columnwidth ]{./imgs/dtu_comparison_mesh/dtu_scan63/neus/24.png}};
\node[inner sep=0pt] (d0) at (\ShiftXLarge*3, \ShiftySmall +\ShiftyLarge*4)
{\includegraphics[ width=.44\columnwidth ]{./imgs/dtu_comparison_mesh/dtu_scan63/colmap/24.png}};
\node[inner sep=0pt] (d0) at (\ShiftXLarge*4, \ShiftySmall +\ShiftyLarge*4)
{\includegraphics[ width=.44\columnwidth ]{./imgs/dtu_comparison_mesh/dtu_scan63/ingp/24.png}};
\node[inner sep=0pt] (d0) at (\ShiftXLarge*1, -\ShiftySmall +\ShiftyLarge*4)
{\includegraphics[ width=.44\columnwidth ]{./imgs/dtu_comparison_error/dtu_scan63/mine/24.png}};
\node[inner sep=0pt] (d0) at (\ShiftXLarge*2, -\ShiftySmall +\ShiftyLarge*4)
{\includegraphics[ width=.44\columnwidth ]{./imgs/dtu_comparison_error/dtu_scan63/neus/24.png}};
\node[inner sep=0pt] (d0) at (\ShiftXLarge*3, -\ShiftySmall +\ShiftyLarge*4)
{\includegraphics[ width=.44\columnwidth ]{./imgs/dtu_comparison_error/dtu_scan63/colmap/24.png}};
\node[inner sep=0pt] (d0) at (\ShiftXLarge*4, -\ShiftySmall +\ShiftyLarge*4)
{\includegraphics[ width=.44\columnwidth ]{./imgs/dtu_comparison_error/dtu_scan63/ingp/24.png}};

\node[] at (\ShiftXLarge*0,2.4) {\footnotesize G.T. point cloud };
\node[] at (\ShiftXLarge*1,2.4) {\footnotesize Ours };
\node[] at (\ShiftXLarge*2,2.4) {\footnotesize NeuS };
\node[] at (\ShiftXLarge*3,2.4) {\footnotesize Colmap };
\node[] at (\ShiftXLarge*4,2.4) {\footnotesize INGP };

\end{tikzpicture}
\vspace{-5mm}
\caption{ DTU qualitative comparison of extracted meshes and error maps.  } \label{fig:dtu1}
\vspace{-3mm}
\end{figure*} %
\begin{figure*}[]
\centering
\small\sffamily
\begin{tikzpicture}

\newcommand\ShiftXLarge{3.0} %
\newcommand\ShiftySmall{1.1} %
\newcommand\ShiftyLarge{-4.5} %

\node[inner sep=0pt] (d0) at (\ShiftXLarge*0, \ShiftyLarge*0)
    {\includegraphics[ width=.38\columnwidth ]{./imgs/dtu_comparison_mesh/dtu_scan65/gt_cloud/23.png}};
\node[inner sep=0pt] (d0) at (\ShiftXLarge*1, \ShiftySmall +\ShiftyLarge*0)
{\includegraphics[ width=.38\columnwidth ]{./imgs/dtu_comparison_mesh/dtu_scan65/mine/23.png}};
\node[inner sep=0pt] (d0) at (\ShiftXLarge*2, \ShiftySmall +\ShiftyLarge*0)
{\includegraphics[ width=.38\columnwidth ]{./imgs/dtu_comparison_mesh/dtu_scan65/neus/23.png}};
\node[inner sep=0pt] (d0) at (\ShiftXLarge*3, \ShiftySmall +\ShiftyLarge*0)
{\includegraphics[ width=.38\columnwidth ]{./imgs/dtu_comparison_mesh/dtu_scan65/colmap/23.png}};
\node[inner sep=0pt] (d0) at (\ShiftXLarge*4, \ShiftySmall +\ShiftyLarge*0)
{\includegraphics[ width=.38\columnwidth ]{./imgs/dtu_comparison_mesh/dtu_scan65/ingp/23.png}};
\node[inner sep=0pt] (d0) at (\ShiftXLarge*1, -\ShiftySmall +\ShiftyLarge*0)
{\includegraphics[ width=.38\columnwidth ]{./imgs/dtu_comparison_error/dtu_scan65/mine/23.png}};
\node[inner sep=0pt] (d0) at (\ShiftXLarge*2, -\ShiftySmall +\ShiftyLarge*0)
{\includegraphics[ width=.38\columnwidth ]{./imgs/dtu_comparison_error/dtu_scan65/neus/23.png}};
\node[inner sep=0pt] (d0) at (\ShiftXLarge*3, -\ShiftySmall +\ShiftyLarge*0)
{\includegraphics[ width=.38\columnwidth ]{./imgs/dtu_comparison_error/dtu_scan65/colmap/23.png}};
\node[inner sep=0pt] (d0) at (\ShiftXLarge*4, -\ShiftySmall +\ShiftyLarge*0)
{\includegraphics[ width=.38\columnwidth ]{./imgs/dtu_comparison_error/dtu_scan65/ingp/23.png}};

\node[inner sep=0pt] (d0) at (\ShiftXLarge*0, \ShiftyLarge*1)
{\includegraphics[ width=.34\columnwidth ]{./imgs/dtu_comparison_mesh/dtu_scan69/gt_cloud/43.png}};
\node[inner sep=0pt] (d0) at (\ShiftXLarge*1, \ShiftySmall +\ShiftyLarge*1)
{\includegraphics[ width=.34\columnwidth ]{./imgs/dtu_comparison_mesh/dtu_scan69/mine/43.png}};
\node[inner sep=0pt] (d0) at (\ShiftXLarge*2, \ShiftySmall +\ShiftyLarge*1)
{\includegraphics[ width=.34\columnwidth ]{./imgs/dtu_comparison_mesh/dtu_scan69/neus/43.png}};
\node[inner sep=0pt] (d0) at (\ShiftXLarge*3, \ShiftySmall +\ShiftyLarge*1)
{\includegraphics[ width=.34\columnwidth ]{./imgs/dtu_comparison_mesh/dtu_scan69/colmap/43.png}};
\node[inner sep=0pt] (d0) at (\ShiftXLarge*4, \ShiftySmall +\ShiftyLarge*1)
{\includegraphics[ width=.34\columnwidth ]{./imgs/dtu_comparison_mesh/dtu_scan69/ingp/43.png}};
\node[inner sep=0pt] (d0) at (\ShiftXLarge*1, -\ShiftySmall +\ShiftyLarge*1)
{\includegraphics[ width=.34\columnwidth ]{./imgs/dtu_comparison_error/dtu_scan69/mine/43.png}};
\node[inner sep=0pt] (d0) at (\ShiftXLarge*2, -\ShiftySmall +\ShiftyLarge*1)
{\includegraphics[ width=.34\columnwidth ]{./imgs/dtu_comparison_error/dtu_scan69/neus/43.png}};
\node[inner sep=0pt] (d0) at (\ShiftXLarge*3, -\ShiftySmall +\ShiftyLarge*1)
{\includegraphics[ width=.34\columnwidth ]{./imgs/dtu_comparison_error/dtu_scan69/colmap/43.png}};
\node[inner sep=0pt] (d0) at (\ShiftXLarge*4, -\ShiftySmall +\ShiftyLarge*1)
{\includegraphics[ width=.34\columnwidth ]{./imgs/dtu_comparison_error/dtu_scan69/ingp/43.png}};

\node[inner sep=0pt] (d0) at (\ShiftXLarge*0, \ShiftyLarge*2)
{\includegraphics[ width=.36\columnwidth ]{./imgs/dtu_comparison_mesh/dtu_scan83/gt_cloud/25.png}};
\node[inner sep=0pt] (d0) at (\ShiftXLarge*1, \ShiftySmall +\ShiftyLarge*2)
{\includegraphics[ width=.36\columnwidth ]{./imgs/dtu_comparison_mesh/dtu_scan83/mine/25.png}};
\node[inner sep=0pt] (d0) at (\ShiftXLarge*2, \ShiftySmall +\ShiftyLarge*2)
{\includegraphics[ width=.36\columnwidth ]{./imgs/dtu_comparison_mesh/dtu_scan83/neus/25.png}};
\node[inner sep=0pt] (d0) at (\ShiftXLarge*3, \ShiftySmall +\ShiftyLarge*2)
{\includegraphics[ width=.36\columnwidth ]{./imgs/dtu_comparison_mesh/dtu_scan83/colmap/25.png}};
\node[inner sep=0pt] (d0) at (\ShiftXLarge*4, \ShiftySmall +\ShiftyLarge*2)
{\includegraphics[ width=.36\columnwidth ]{./imgs/dtu_comparison_mesh/dtu_scan83/ingp/25.png}};
\node[inner sep=0pt] (d0) at (\ShiftXLarge*1, -\ShiftySmall +\ShiftyLarge*2)
{\includegraphics[ width=.36\columnwidth ]{./imgs/dtu_comparison_error/dtu_scan83/mine/25.png}};
\node[inner sep=0pt] (d0) at (\ShiftXLarge*2, -\ShiftySmall +\ShiftyLarge*2)
{\includegraphics[ width=.36\columnwidth ]{./imgs/dtu_comparison_error/dtu_scan83/neus/25.png}};
\node[inner sep=0pt] (d0) at (\ShiftXLarge*3, -\ShiftySmall +\ShiftyLarge*2)
{\includegraphics[ width=.36\columnwidth ]{./imgs/dtu_comparison_error/dtu_scan83/colmap/25.png}};
\node[inner sep=0pt] (d0) at (\ShiftXLarge*4, -\ShiftySmall +\ShiftyLarge*2)
{\includegraphics[ width=.36\columnwidth ]{./imgs/dtu_comparison_error/dtu_scan83/ingp/25.png}};

\node[inner sep=0pt] (d0) at (\ShiftXLarge*0, \ShiftyLarge*3)
{\includegraphics[ width=.34\columnwidth ]{./imgs/dtu_comparison_mesh/dtu_scan97/gt_cloud/24.png}};
\node[inner sep=0pt] (d0) at (\ShiftXLarge*1, \ShiftySmall +\ShiftyLarge*3)
{\includegraphics[ width=.34\columnwidth ]{./imgs/dtu_comparison_mesh/dtu_scan97/mine/24.png}};
\node[inner sep=0pt] (d0) at (\ShiftXLarge*2, \ShiftySmall +\ShiftyLarge*3)
{\includegraphics[ width=.34\columnwidth ]{./imgs/dtu_comparison_mesh/dtu_scan97/neus/24.png}};
\node[inner sep=0pt] (d0) at (\ShiftXLarge*3, \ShiftySmall +\ShiftyLarge*3)
{\includegraphics[ width=.34\columnwidth ]{./imgs/dtu_comparison_mesh/dtu_scan97/colmap/24.png}};
\node[inner sep=0pt] (d0) at (\ShiftXLarge*4, \ShiftySmall +\ShiftyLarge*3)
{\includegraphics[ width=.34\columnwidth ]{./imgs/dtu_comparison_mesh/dtu_scan97/ingp/24.png}};
\node[inner sep=0pt] (d0) at (\ShiftXLarge*1, -\ShiftySmall +\ShiftyLarge*3)
{\includegraphics[ width=.34\columnwidth ]{./imgs/dtu_comparison_error/dtu_scan97/mine/24.png}};
\node[inner sep=0pt] (d0) at (\ShiftXLarge*2, -\ShiftySmall +\ShiftyLarge*3)
{\includegraphics[ width=.34\columnwidth ]{./imgs/dtu_comparison_error/dtu_scan97/neus/24.png}};
\node[inner sep=0pt] (d0) at (\ShiftXLarge*3, -\ShiftySmall +\ShiftyLarge*3)
{\includegraphics[ width=.34\columnwidth ]{./imgs/dtu_comparison_error/dtu_scan97/colmap/24.png}};
\node[inner sep=0pt] (d0) at (\ShiftXLarge*4, -\ShiftySmall +\ShiftyLarge*3)
{\includegraphics[ width=.34\columnwidth ]{./imgs/dtu_comparison_error/dtu_scan97/ingp/24.png}};

\node[inner sep=0pt] (d0) at (\ShiftXLarge*0, \ShiftyLarge*4)
{\includegraphics[ width=.34\columnwidth ]{./imgs/dtu_comparison_mesh/dtu_scan105/gt_cloud/13.png}};
\node[inner sep=0pt] (d0) at (\ShiftXLarge*1, \ShiftySmall +\ShiftyLarge*4)
{\includegraphics[ width=.34\columnwidth ]{./imgs/dtu_comparison_mesh/dtu_scan105/mine/13.png}};
\node[inner sep=0pt] (d0) at (\ShiftXLarge*2, \ShiftySmall +\ShiftyLarge*4)
{\includegraphics[ width=.34\columnwidth ]{./imgs/dtu_comparison_mesh/dtu_scan105/neus/13.png}};
\node[inner sep=0pt] (d0) at (\ShiftXLarge*3, \ShiftySmall +\ShiftyLarge*4)
{\includegraphics[ width=.34\columnwidth ]{./imgs/dtu_comparison_mesh/dtu_scan105/colmap/13.png}};
\node[inner sep=0pt] (d0) at (\ShiftXLarge*4, \ShiftySmall +\ShiftyLarge*4)
{\includegraphics[ width=.34\columnwidth ]{./imgs/dtu_comparison_mesh/dtu_scan105/ingp/13.png}};
\node[inner sep=0pt] (d0) at (\ShiftXLarge*1, -\ShiftySmall +\ShiftyLarge*4)
{\includegraphics[ width=.34\columnwidth ]{./imgs/dtu_comparison_error/dtu_scan105/mine/13.png}};
\node[inner sep=0pt] (d0) at (\ShiftXLarge*2, -\ShiftySmall +\ShiftyLarge*4)
{\includegraphics[ width=.34\columnwidth ]{./imgs/dtu_comparison_error/dtu_scan105/neus/13.png}};
\node[inner sep=0pt] (d0) at (\ShiftXLarge*3, -\ShiftySmall +\ShiftyLarge*4)
{\includegraphics[ width=.34\columnwidth ]{./imgs/dtu_comparison_error/dtu_scan105/colmap/13.png}};
\node[inner sep=0pt] (d0) at (\ShiftXLarge*4, -\ShiftySmall +\ShiftyLarge*4)
{\includegraphics[ width=.34\columnwidth ]{./imgs/dtu_comparison_error/dtu_scan105/ingp/13.png}};

\node[] at (\ShiftXLarge*0,2.4) {\footnotesize G.T. point cloud };
\node[] at (\ShiftXLarge*1,2.4) {\footnotesize Ours };
\node[] at (\ShiftXLarge*2,2.4) {\footnotesize NeuS };
\node[] at (\ShiftXLarge*3,2.4) {\footnotesize Colmap };
\node[] at (\ShiftXLarge*4,2.4) {\footnotesize INGP };

\end{tikzpicture}
\vspace{-2mm}
\caption{ DTU qualitative comparison of extracted meshes and error maps.   } \label{fig:dtu2}
\vspace{-3mm}
\end{figure*} %
\begin{figure*}[]
\centering
\small\sffamily
\begin{tikzpicture}

\newcommand\ShiftXLarge{3.0} %
\newcommand\ShiftySmall{1.1} %
\newcommand\ShiftyLarge{-4.5} %

\node[inner sep=0pt] (d0) at (\ShiftXLarge*0, \ShiftyLarge*0)
    {\includegraphics[ width=.42\columnwidth ]{./imgs/dtu_comparison_mesh/dtu_scan106/gt_cloud/51.png}};
\node[inner sep=0pt] (d0) at (\ShiftXLarge*1, \ShiftySmall +\ShiftyLarge*0)
{\includegraphics[ width=.42\columnwidth ]{./imgs/dtu_comparison_mesh/dtu_scan106/mine/51.png}};
\node[inner sep=0pt] (d0) at (\ShiftXLarge*2, \ShiftySmall +\ShiftyLarge*0)
{\includegraphics[ width=.42\columnwidth ]{./imgs/dtu_comparison_mesh/dtu_scan106/neus/51.png}};
\node[inner sep=0pt] (d0) at (\ShiftXLarge*3, \ShiftySmall +\ShiftyLarge*0)
{\includegraphics[ width=.42\columnwidth ]{./imgs/dtu_comparison_mesh/dtu_scan106/colmap/51.png}};
\node[inner sep=0pt] (d0) at (\ShiftXLarge*4, \ShiftySmall +\ShiftyLarge*0)
{\includegraphics[ width=.42\columnwidth ]{./imgs/dtu_comparison_mesh/dtu_scan106/ingp/51.png}};
\node[inner sep=0pt] (d0) at (\ShiftXLarge*1, -\ShiftySmall +\ShiftyLarge*0)
{\includegraphics[ width=.42\columnwidth ]{./imgs/dtu_comparison_error/dtu_scan106/mine/51.png}};
\node[inner sep=0pt] (d0) at (\ShiftXLarge*2, -\ShiftySmall +\ShiftyLarge*0)
{\includegraphics[ width=.42\columnwidth ]{./imgs/dtu_comparison_error/dtu_scan106/neus/51.png}};
\node[inner sep=0pt] (d0) at (\ShiftXLarge*3, -\ShiftySmall +\ShiftyLarge*0)
{\includegraphics[ width=.42\columnwidth ]{./imgs/dtu_comparison_error/dtu_scan106/colmap/51.png}};
\node[inner sep=0pt] (d0) at (\ShiftXLarge*4, -\ShiftySmall +\ShiftyLarge*0)
{\includegraphics[ width=.42\columnwidth ]{./imgs/dtu_comparison_error/dtu_scan106/ingp/51.png}};

\node[inner sep=0pt] (d0) at (\ShiftXLarge*0, \ShiftyLarge*1)
{\includegraphics[ width=.42\columnwidth ]{./imgs/dtu_comparison_mesh/dtu_scan110/gt_cloud/49.png}};
\node[inner sep=0pt] (d0) at (\ShiftXLarge*1, \ShiftySmall +\ShiftyLarge*1)
{\includegraphics[ width=.42\columnwidth ]{./imgs/dtu_comparison_mesh/dtu_scan110/mine/49.png}};
\node[inner sep=0pt] (d0) at (\ShiftXLarge*2, \ShiftySmall +\ShiftyLarge*1)
{\includegraphics[ width=.42\columnwidth ]{./imgs/dtu_comparison_mesh/dtu_scan110/neus/49.png}};
\node[inner sep=0pt] (d0) at (\ShiftXLarge*3, \ShiftySmall +\ShiftyLarge*1)
{\includegraphics[ width=.42\columnwidth ]{./imgs/dtu_comparison_mesh/dtu_scan110/colmap/49.png}};
\node[inner sep=0pt] (d0) at (\ShiftXLarge*4, \ShiftySmall +\ShiftyLarge*1)
{\includegraphics[ width=.42\columnwidth ]{./imgs/dtu_comparison_mesh/dtu_scan110/ingp/49.png}};
\node[inner sep=0pt] (d0) at (\ShiftXLarge*1, -\ShiftySmall +\ShiftyLarge*1)
{\includegraphics[ width=.42\columnwidth ]{./imgs/dtu_comparison_error/dtu_scan110/mine/49.png}};
\node[inner sep=0pt] (d0) at (\ShiftXLarge*2, -\ShiftySmall +\ShiftyLarge*1)
{\includegraphics[ width=.42\columnwidth ]{./imgs/dtu_comparison_error/dtu_scan110/neus/49.png}};
\node[inner sep=0pt] (d0) at (\ShiftXLarge*3, -\ShiftySmall +\ShiftyLarge*1)
{\includegraphics[ width=.42\columnwidth ]{./imgs/dtu_comparison_error/dtu_scan110/colmap/49.png}};
\node[inner sep=0pt] (d0) at (\ShiftXLarge*4, -\ShiftySmall +\ShiftyLarge*1)
{\includegraphics[ width=.42\columnwidth ]{./imgs/dtu_comparison_error/dtu_scan110/ingp/49.png}};

\node[inner sep=0pt] (d0) at (\ShiftXLarge*0, \ShiftyLarge*2)
{\includegraphics[ width=.34\columnwidth ]{./imgs/dtu_comparison_mesh/dtu_scan114/gt_cloud/31.png}};
\node[inner sep=0pt] (d0) at (\ShiftXLarge*1, \ShiftySmall +\ShiftyLarge*2)
{\includegraphics[ width=.34\columnwidth ]{./imgs/dtu_comparison_mesh/dtu_scan114/mine/31.png}};
\node[inner sep=0pt] (d0) at (\ShiftXLarge*2, \ShiftySmall +\ShiftyLarge*2)
{\includegraphics[ width=.34\columnwidth ]{./imgs/dtu_comparison_mesh/dtu_scan114/neus/31.png}};
\node[inner sep=0pt] (d0) at (\ShiftXLarge*3, \ShiftySmall +\ShiftyLarge*2)
{\includegraphics[ width=.34\columnwidth ]{./imgs/dtu_comparison_mesh/dtu_scan114/colmap/31.png}};
\node[inner sep=0pt] (d0) at (\ShiftXLarge*4, \ShiftySmall +\ShiftyLarge*2)
{\includegraphics[ width=.34\columnwidth ]{./imgs/dtu_comparison_mesh/dtu_scan114/ingp/31.png}};
\node[inner sep=0pt] (d0) at (\ShiftXLarge*1, -\ShiftySmall +\ShiftyLarge*2)
{\includegraphics[ width=.34\columnwidth ]{./imgs/dtu_comparison_error/dtu_scan114/mine/31.png}};
\node[inner sep=0pt] (d0) at (\ShiftXLarge*2, -\ShiftySmall +\ShiftyLarge*2)
{\includegraphics[ width=.34\columnwidth ]{./imgs/dtu_comparison_error/dtu_scan114/neus/31.png}};
\node[inner sep=0pt] (d0) at (\ShiftXLarge*3, -\ShiftySmall +\ShiftyLarge*2)
{\includegraphics[ width=.34\columnwidth ]{./imgs/dtu_comparison_error/dtu_scan114/colmap/31.png}};
\node[inner sep=0pt] (d0) at (\ShiftXLarge*4, -\ShiftySmall +\ShiftyLarge*2)
{\includegraphics[ width=.34\columnwidth ]{./imgs/dtu_comparison_error/dtu_scan114/ingp/31.png}};

\node[inner sep=0pt] (d0) at (\ShiftXLarge*0, \ShiftyLarge*3)
{\includegraphics[ width=.39\columnwidth ]{./imgs/dtu_comparison_mesh/dtu_scan118/gt_cloud/60.png}};
\node[inner sep=0pt] (d0) at (\ShiftXLarge*1, \ShiftySmall +\ShiftyLarge*3)
{\includegraphics[ width=.39\columnwidth ]{./imgs/dtu_comparison_mesh/dtu_scan118/mine/60.png}};
\node[inner sep=0pt] (d0) at (\ShiftXLarge*2, \ShiftySmall +\ShiftyLarge*3)
{\includegraphics[ width=.39\columnwidth ]{./imgs/dtu_comparison_mesh/dtu_scan118/neus/60.png}};
\node[inner sep=0pt] (d0) at (\ShiftXLarge*3, \ShiftySmall +\ShiftyLarge*3)
{\includegraphics[ width=.39\columnwidth ]{./imgs/dtu_comparison_mesh/dtu_scan118/colmap/60.png}};
\node[inner sep=0pt] (d0) at (\ShiftXLarge*4, \ShiftySmall +\ShiftyLarge*3)
{\includegraphics[ width=.39\columnwidth ]{./imgs/dtu_comparison_mesh/dtu_scan118/ingp/60.png}};
\node[inner sep=0pt] (d0) at (\ShiftXLarge*1, -\ShiftySmall +\ShiftyLarge*3)
{\includegraphics[ width=.39\columnwidth ]{./imgs/dtu_comparison_error/dtu_scan118/mine/60.png}};
\node[inner sep=0pt] (d0) at (\ShiftXLarge*2, -\ShiftySmall +\ShiftyLarge*3)
{\includegraphics[ width=.39\columnwidth ]{./imgs/dtu_comparison_error/dtu_scan118/neus/60.png}};
\node[inner sep=0pt] (d0) at (\ShiftXLarge*3, -\ShiftySmall +\ShiftyLarge*3)
{\includegraphics[ width=.39\columnwidth ]{./imgs/dtu_comparison_error/dtu_scan118/colmap/60.png}};
\node[inner sep=0pt] (d0) at (\ShiftXLarge*4, -\ShiftySmall +\ShiftyLarge*3)
{\includegraphics[ width=.39\columnwidth ]{./imgs/dtu_comparison_error/dtu_scan118/ingp/60.png}};

\node[inner sep=0pt] (d0) at (\ShiftXLarge*0, \ShiftyLarge*4)
{\includegraphics[ width=.39\columnwidth ]{./imgs/dtu_comparison_mesh/dtu_scan122/gt_cloud/55.png}};
\node[inner sep=0pt] (d0) at (\ShiftXLarge*1, \ShiftySmall +\ShiftyLarge*4)
{\includegraphics[ width=.39\columnwidth ]{./imgs/dtu_comparison_mesh/dtu_scan122/mine/55.png}};
\node[inner sep=0pt] (d0) at (\ShiftXLarge*2, \ShiftySmall +\ShiftyLarge*4)
{\includegraphics[ width=.39\columnwidth ]{./imgs/dtu_comparison_mesh/dtu_scan122/neus/55.png}};
\node[inner sep=0pt] (d0) at (\ShiftXLarge*3, \ShiftySmall +\ShiftyLarge*4)
{\includegraphics[ width=.39\columnwidth ]{./imgs/dtu_comparison_mesh/dtu_scan122/colmap/55.png}};
\node[inner sep=0pt] (d0) at (\ShiftXLarge*4, \ShiftySmall +\ShiftyLarge*4)
{\includegraphics[ width=.39\columnwidth ]{./imgs/dtu_comparison_mesh/dtu_scan122/ingp/55.png}};
\node[inner sep=0pt] (d0) at (\ShiftXLarge*1, -\ShiftySmall +\ShiftyLarge*4)
{\includegraphics[ width=.39\columnwidth ]{./imgs/dtu_comparison_error/dtu_scan122/mine/55.png}};
\node[inner sep=0pt] (d0) at (\ShiftXLarge*2, -\ShiftySmall +\ShiftyLarge*4)
{\includegraphics[ width=.39\columnwidth ]{./imgs/dtu_comparison_error/dtu_scan122/neus/55.png}};
\node[inner sep=0pt] (d0) at (\ShiftXLarge*3, -\ShiftySmall +\ShiftyLarge*4)
{\includegraphics[ width=.39\columnwidth ]{./imgs/dtu_comparison_error/dtu_scan122/colmap/55.png}};
\node[inner sep=0pt] (d0) at (\ShiftXLarge*4, -\ShiftySmall +\ShiftyLarge*4)
{\includegraphics[ width=.39\columnwidth ]{./imgs/dtu_comparison_error/dtu_scan122/ingp/55.png}};

\node[] at (\ShiftXLarge*0,2.4) {\footnotesize G.T. point cloud };
\node[] at (\ShiftXLarge*1,2.4) {\footnotesize Ours };
\node[] at (\ShiftXLarge*2,2.4) {\footnotesize NeuS };
\node[] at (\ShiftXLarge*3,2.4) {\footnotesize Colmap };
\node[] at (\ShiftXLarge*4,2.4) {\footnotesize INGP };

\end{tikzpicture}
\vspace{-5mm}
\caption{ DTU qualitative comparison of extracted meshes and error maps.   } \label{fig:dtu3}
\vspace{-3mm}
\end{figure*} 
In~\reffig{fig:dtu1} -- \reffig{fig:dtu3}, we show additional qualitative results from the DTU dataset~\cite{dtu}. We show extracted meshes and error maps which represent the distance from each mesh vertex towards the nearest point from the ground-truth. Please note that the ground-truth can have holes in areas of high reflectance or self-occlusion and this shows as a bright yellow color in the error map.

\end{document}